%% file: main.tex
\pdfoutput=1
\documentclass{article}
\usepackage{silence}
\usepackage{microtype}
\usepackage{graphicx}
\usepackage{booktabs} %
\usepackage{amsmath}
\usepackage{amssymb}  %
\usepackage{xcolor}
\usepackage{tikz}
\definecolor{stepink}{HTML}{1A202C}
\newcommand{\stepnum}[1]{\tikz[baseline=(n.base)]\node[circle,fill=stepink,text=white,inner sep=0pt,minimum size=1.2em,font=\scriptsize\sffamily\bfseries](n){#1};} %

\usetikzlibrary{patterns,arrows.meta} %
\usepackage{pgfplots}
\pgfplotsset{compat=1.17}

\usepackage{float}
\usepackage[hyperfootnotes=false]{hyperref} %
\usepackage{xurl} %
\usepackage{enumitem}
\setlist[itemize]{leftmargin=*,topsep=2pt,itemsep=1pt,parsep=0pt} %

\def\submissionbuild{1}
\usepackage[accepted]{mlsys2025}
\makeatletter\renewcommand{\Notice@String}{}\makeatother
\AtBeginDocument{\hypersetup{pdfsubject={}}}

\ifdefined\submissionbuild

\newcommand{\rev}[1]{#1}
\newcommand{\revorange}[1]{#1}
\pdfinfoomitdate=1 \pdftrailerid{} %
\else

\newcommand{\rev}[1]{\textcolor{red}{#1}}
\newcommand{\revorange}[1]{\textcolor{orange}{#1}} %
\fi
\makeatletter
\newif\if@apxpart
\newcommand\apxlabel[1]{\expandafter\gdef\csname apx@#1\endcsname{}}
\AddToHookWithArguments{label}{\if@apxpart\protected@write\@auxout{}{\string\apxlabel{#1}}\fi}
\newcommand\apxstartpart{%
  \global\@apxparttrue
  \renewcommand*{\hyper@natlinkstart}[1]{}%
  \renewcommand*{\hyper@natlinkend}{}%
  \renewcommand*{\hyper@natlinkbreak}[2]{##1}}
\AtBeginDocument{%
  \let\apx@Tref\T@ref
  \def\T@ref#1{%
    \@ifundefined{apx@#1}%
      {\if@apxpart\apx@plain{#1}\else\apx@Tref{#1}\fi}%
      {\if@apxpart\apx@Tref{#1}\else\apx@plain{#1}\fi}}}
\newcommand\apx@plain[1]{{\color{black}\@refstar{#1}}}
\makeatother

\input{numbers}

\mlsystitlerunning{How Much Mismatch Can On-Policy Distillation Tolerate?}

\begin{document}

\twocolumn[
\mlsystitle{How Much Training--Inference Mismatch Can On-Policy Distillation Tolerate?}

\mlsyssetsymbol{equal}{*}

\begin{mlsysauthorlist}
\mlsysauthor{Cong Yu}{aalto}
\mlsysauthor{Ziting Wang}{ntu}
\mlsysauthor{Songlin Jiang}{aalto}
\mlsysauthor{Sitong Zhang}{indep}
\\
\mlsysauthor{Chengliang Chai}{bit}
\mlsysauthor{Youhe Jiang}{cam}
\mlsysauthor{Eiko Yoneki}{cam}
\mlsysauthor{Tuo Shi}{suat}
\end{mlsysauthorlist}

{\centering\small \mbox{\textsuperscript{1}Aalto University} \quad \mbox{\textsuperscript{2}Nanyang Technological University} \quad \mbox{\textsuperscript{3}Independent} \quad \mbox{\textsuperscript{4}Beijing Institute of Technology} \quad \mbox{\textsuperscript{5}University of Cambridge} \quad \mbox{\textsuperscript{6}Shenzhen University of Advanced Technology}\par}

\mlsyskeywords{on-policy distillation, training--inference mismatch, low-precision rollouts, quantization, asynchronous rollouts, reasoning LLMs}

\vskip 0.3in

\input{sections/00-abstract}
]

\input{sections/n01-introduction}

\input{sections/n02-background}
\input{sections/s03-overview}
\input{sections/s04-setup}

\input{sections/s05-evaluation}

\input{sections/n06-discussion}

\input{sections/n07-related}

\input{sections/n08-conclusion}

\bibliography{main}
\bibliographystyle{mlsys2025}

\clearpage
\twocolumn[{\centering{\large\bfseries Appendix\par}
\vskip 1.2em\label{appendix-start}}]
\input{sections/a-about} %
\input{sections/a-prereg}     %
\input{sections/a-engines}    %
\input{sections/a-probe}      %
\input{sections/a-speedrepro} %
\input{sections/a-tables}
\input{sections/a-limitations} %

\end{document}

%% file: numbers.tex
\newcommand{\ProbeRows}{442{,}792}
\newcommand{\ProbeTraces}{64}
\newcommand{\ProbeRbiasNmid}{0.079}
\newcommand{\ProbeRbiasPctNmid}{8}
\newcommand{\ProbeBiasProjNmid}{\ensuremath{5.83\times10^{-5}}}
\newcommand{\ProbeMMNmid}{0.0086}
\newcommand{\ProbeEqANmid}{\ensuremath{-0.0705}}
\newcommand{\ProbeRbiasNhigh}{0.267}
\newcommand{\ProbeRbiasPctNhigh}{27}
\newcommand{\ProbeBiasProjNhigh}{\ensuremath{1.49\times10^{-4}}}
\newcommand{\ProbeMMNhigh}{0.044}
\newcommand{\ProbeEqANhigh}{\ensuremath{-0.0904}}
\newcommand{\ProbeIdentityErr}{\ensuremath{5.2\times10^{-5}}}
\newcommand{\StepSCleanMean}{107.5}
\newcommand{\TimeGenCleanPct}{68}
\newcommand{\AmdahlGzerob}{1.16}
\newcommand{\AmdahlCeiling}{3.08}
\newcommand{\GzeroBest}{1.25 / 1.10 / 1.19}
\newcommand{\GzeroThr}{1.43 / 1.37 / 1.38}
\newcommand{\GzeroKVcapKV}{1.93}
\newcommand{\GzeroMismatch}{0.0043 [0.0040, 0.0046]}
\newcommand{\GzeroKVuseMax}{0.71}
\newcommand{\GzeroPreemptMax}{0}
\newcommand{\ProbeAngleNmid}{4.2}
\newcommand{\ProbeAngleNhigh}{12.2}

%% file: sections/00-abstract.tex
\begin{abstract}
On-policy distillation (OPD) trains a student language model on its own responses, guided by a fixed teacher.
Generating these responses is expensive, so systems use quantized inference and asynchronous rollouts to reduce the cost. 
Both, however, introduce training–inference mismatch between the rollout engine and the model being trained.
Such mismatch can destabilize reinforcement learning (RL), but its effects on OPD remain unclear. 
This paper investigates how much mismatch OPD tolerates, where its effects arise, and how training systems can use this tolerance. 
\revorange{Measuring all mismatch sources on one scale, we examine} quantization across three student–teacher pairs, with additional tests of weight staleness and injected noise on one pair. OPD tolerates substantial mismatch without meaningful loss of training quality but degrades at higher levels, revealing an empirical mismatch budget. Both \revorange{the quality cost of mismatch} and the mismatch introduced by quantization depend on the model, with some layers particularly sensitive to reduced precision. In controlled comparisons, OPD remains stable under rollout configurations that destabilize RL methods. Under weight noise, mismatch also shifts the prefixes reached during generation, introducing errors that per-token corrections cannot address. To put these findings into practice, we introduce a lightweight admission check that screens rollout configurations against a calibrated budget without requiring a full training run for each candidate. Model-specific FP8 recipes preserve training quality while providing capacity for at least 1.93× as many KV-cache tokens as BF16. Combining FP8 with asynchronous rollouts reaches the same \revorange{training quality} 1.62× sooner than synchronous BF16 training.
Code is available at \url{https://github.com/benyucong/train-inference-mismatch-opd/}.

\end{abstract}

%% file: sections/n01-introduction.tex
\section{Introduction}
\label{sec:intro}

On-policy distillation (OPD) trains a student model on responses it samples itself, guided by a fixed teacher that scores every token \citep{agarwal2024gkd,gu2024minillm,lu2025onpolicy}. 
It brings a stronger or specialized teacher's ability into the student, and many recent model releases run it as a post-training stage \citep{xiaomi2026mimo,yang2026nemotroncascade2,nvidia2026nemotron3ultra}. 
Its cost lies in producing the responses, called \emph{rollouts}, which can dominate training time \citep{kang2026asyncopd,qiu2026fp8rl}. 
\revorange{With a standard inference engine in 16-bit BF16, rollout generation and teacher scoring take 76\% of each training step's time on the 7B student (\revorange{App.}~Table~\ref{tab:speed-64}).} Therefore, cheaper rollouts mean cheaper OPD.

Two techniques make rollouts cheaper. 
\emph{Quantized inference} stores the rollout engine's weights and key-value (KV) cache in fewer bits, such as the 8-bit FP8 format, so the same GPUs hold more rollouts \citep{qiu2026fp8rl,liu2025flashrl}. 
\emph{Asynchronous rollouts} let generation run ahead of training, so the trainer does not wait for it, and the rollouts come from weights a few versions old \citep{fu2025areal,kang2026asyncopd}. 
However, both come at a price: the \emph{rollout engine} that generates the rollouts no longer runs exactly the model being trained, so the student learns from responses that are not quite its own. 
We call this gap \emph{mismatch} and measure it as the per-token KL divergence between the two models' next-token distributions, in nats per token. 
Even a standard engine has a small mismatch, because inference engines and trainers use different kernels and numerics \citep{liu2025speedkills,he2025nondeterminism}, and quantization and stale weights widen it further.

\revorange{RL post-training can face the same training--inference mismatch. It also trains a model on its own rollouts, and the mismatch can collapse training} \citep{zhong2026vexact}.
\revorange{Many RL systems therefore correct the update per token} \citep{yao2025offpolicy,marek2026score}.
\revorange{Recent OPD pipelines often apply these remedies} \citep{xiaomi2026mimo,yang2026nemotroncascade2,nvidia2026nemotron3ultra}.

\begin{figure}[!t]
\centering
\includegraphics[width=\columnwidth]{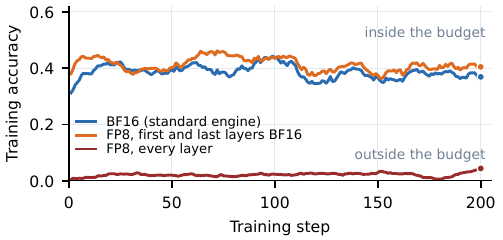}
\caption{\textbf{OPD tolerates the mismatch of quantized rollouts, up to a limit.} \revorange{Training accuracy (the
fraction of the rollout engine's responses whose final answer is correct, smoothed over 21 steps) for OPD on the 7B
student, one training run per line.}}
\label{fig:intro-accuracy-time}
\end{figure} 
However, OPD appears far less sensitive to mismatch than RL. 
Semi-OPD \citep{zhao2026semiopd} and Lightning OPD \citep{wu2026lightning} generate rollouts once, from the untrained student, and distill on them throughout training, so the student drifts \revorange{away} from the rollouts. 
They still match or beat OPD with fresh rollouts in many settings. AsyncOPD \citep{kang2026asyncopd} lets generation run many batches ahead of training \revorange{and} still reaches the accuracy of synchronous training. \looseness=-1

\revorange{We call this tolerance the \emph{mismatch budget}: the mismatch OPD can absorb without hurting training, which quantized and asynchronous rollouts can spend on memory and speed.
We judge training by \emph{held-out KL}, the student's per-token KL to the teacher on held-out problems (\S\ref{sec:budget-method}), with accuracy as a second check.
Figure~\ref{fig:intro-accuracy-time} shows \revorange{training accuracy} on a 7B student. 
Inside the budget, \revorange{FP8 rollouts that keep the first and last layers in BF16 track the standard engine step for step, and rollouts up to 4 versions old train as well as it and finish up to 1.6 times sooner (\S\ref{sec:speed}).} 
Beyond it, every-layer FP8 rollouts are rarely correct and training fails \revorange{(\S\ref{sec:budget})}. 
Whether an engine is safe is decided by the budget, not by its bit width (\S\ref{sec:shortcuts}): on this student, only quantized engines that keep the first layer's KV cache in BF16 stay inside the budget.}

Measuring this budget and leveraging it is challenging for three reasons. 
\textbf{(1)}~\emph{Different sources of mismatch are hard to compare.} Quantization is set in bits and asynchrony in versions, and prior OPD studies keep the engine fixed or study staleness alone \citep{zhao2026semiopd,wu2026lightning, kang2026asyncopd,chen2026fopd}, so no common scale relates the two. 
\textbf{(2)}~\emph{Checking an engine costs a training run.} \revorange{A run takes 5.5 to 6 hours on eight GPUs (\S\ref{sec:shortcuts}), too long to try engines one by one; a useful check should take minutes.}
\textbf{(3)}~\emph{Corrections inherited from RL may miss the damage.} \revorange{\rev{They act only on the error at each sampled token. Whether that is where mismatch hurts OPD, and} whether these corrections widen the budget\rev{,} has not been tested.} 

We address these challenges through a measurement study.
First, we establish a common mismatch measure to compare
different rollout engines, define the mismatch budget,
and develop a lightweight admission check to evaluate
new configurations without full training runs
(\S\ref{sec:overview}).
Using controlled rollout configurations and evaluation
criteria (\S\ref{sec:setup}), we characterize OPD's
mismatch tolerance across student models, validate
the admission check, and quantify the efficiency gains
of cheaper rollouts (\S\ref{sec:eval}).
Finally, we investigate why OPD behaves differently
from RL under mismatch and analyze the errors that
per-token corrections may miss (\S\ref{sec:discussion}).

The study covers three student--teacher pairs from two model families, with 1.5B, 3B and 7B students. \revorange{Each
student is distilled for 200 steps from an RL-fine-tuned version of itself. On the 7B pair, we train 17 rollout
configurations (\S\ref{sec:setup-samplers}): six quantized ones from three production recipes, five with
different degrees of asynchrony, four with different levels of weight noise, and two combining quantization with
asynchrony; the 1.5B and 3B pairs add three and two FP8 runs.}
Our contributions are as follows:
\begin{itemize}
\setlength{\itemsep}{1pt}\setlength{\parskip}{0pt}

\item \textbf{A methodology for quantifying OPD's
mismatch budget} (\S\ref{sec:measure}, \S\ref{sec:budget-method}, \S\ref{sec:budget}, \S\ref{sec:students}).
We define a \revorange{common mismatch measure and a procedure for finding}
OPD's mismatch budget. On the 7B student, OPD
preserves training quality at over 200 times the
standard engine's mismatch \revorange{and fails}
at about twice that level\revorange{; under the same engines, two RL methods
break at about 12 and 60 times}. \revorange{Both}
mismatch and its effects vary across \revorange{students}.

\item \textbf{An admission check that takes minutes} (\S\ref{sec:admission-overview},
\S\ref{sec:shortcuts}, \S\ref{sec:speed}). \revorange{Decisions fixed before training matched
all eight tested outcomes, five from the offline check,
checking six configurations in 24 minutes on one GPU.} Within the
budget, FP8 KV caches provide at least
\revorange{1.93 times BF16's capacity}, while FP8 with asynchrony
reaches \revorange{synchronous training's} final held-out KL
\revorange{1.62 times sooner}.

\item \textbf{An analysis of OPD's tolerance and its limits} (\S\ref{sec:rl-contrast} and
\S\ref{sec:discussion}).
We analyze why OPD tolerates mismatch differently
from RL. Mismatch perturbs every OPD update but
leaves its goal of matching the teacher unchanged.
Its impact grows late in training, and runs beyond
the observed budget fail. Under configurations OPD
tolerates, uncorrected RL breaks in five of six runs.
Per-token corrections can prevent these RL failures
but do not improve OPD: the prefix-shift error they
cannot address grows 5.3 times from a passing to a
failing noise level, versus 2.1 times for per-token
error.

\end{itemize}

%% file: sections/n02-background.tex
\section{Background and Motivation}
\label{sec:background}

\noindent
\textbf{On-policy post-training.}
RL post-training and OPD are both \emph{on-policy}: the model learns from responses it samples itself, and both use
the same update. The model $p_\theta$ samples a response $y$ to a prompt. Each token $y_t$ follows a partial response
$y_{<t}$, its \emph{prefix}, and receives a score $A_t$, its \emph{advantage}. The update
$
  \textstyle\sum_t A_t\,\nabla_\theta\log p_\theta(y_t\mid y_{<t})
$
raises the probability of tokens with a positive advantage and lowers it for those with a negative one, so the
advantage decides what the model learns.

The two methods differ only in where the advantage comes from. In RL
\citep{williams1992reinforce,schulman2017ppo,ahmadian2024back,shao2024deepseekmath,guo2025deepseekr1,yu2025dapo}, it
comes from a reward on the whole response, such as whether its final answer is correct, so every token of a response
shares one score. OPD \citep{agarwal2024gkd,gu2024minillm,lu2025onpolicy} instead lets a fixed teacher $p_T$ score
every token:
\begin{equation}\label{eq:bg-adv}
  A_t=\log p_T(y_t\mid y_{<t})-\log p_\theta(y_t\mid y_{<t}),
\end{equation}
which is positive when the teacher finds the token likelier than the student does. \revorange{Figure~\ref{fig:overview}
shows one training step.}

\revorange{This difference decides where the update vanishes. In RL, a correct response keeps its reward no matter how
good the model already is, so each update pushes up the correct responses it sampled. These pushes cancel on average
only if responses are sampled exactly as the model would sample them \citep{williams1992reinforce}; otherwise they
leave a net push. In OPD, the advantage is zero at every token once the student matches the teacher, so the update
vanishes there, whichever responses it trains on. Before that point, averaged over the student's own samples, the
update lowers $\mathrm{KL}(p_\theta\,\|\,p_T)$.} \looseness=-1

\textbf{Rollouts and mismatch.}
Rollout generation is a major cost of on-policy training \citep{kang2026asyncopd,qiu2026fp8rl}, so systems use a
rollout engine such as vLLM \citep{kwon2023pagedattention}. Even with synchronized weights, the engine and trainer
can produce slightly different probabilities due to different kernels and numerics
\citep{liu2025speedkills,he2025nondeterminism}. Two speedups widen this mismatch.
\textbf{(i)}~Quantization stores engine weights or KV cache in 8 or 4 bits to fit more responses in memory, while
rounding further shifts probabilities \citep{liu2025flashrl,li2026qurl}.
\textbf{(ii)}~Asynchrony overlaps generation and training, so the engine uses weights one or more updates old
\citep{fu2025areal,rang2026nearpolicy,kang2026asyncopd}.

Training--inference mismatch also affects RL post-training, where its consequences are better understood. In RL, this
mismatch can break training. RL's update is unbiased only over the model's own samples, so the engine's samples skew
it regardless of reward. Each weight sync copies \revorange{this drift} back into the engine, where it can accumulate
\citep{marek2026score}. Even a standard vLLM engine's small mismatch can make RL training fail, while an exactly
matched engine does not \citep{zhong2026vexact}.

\textbf{Handling mismatch in RL.}
RL handles mismatch mainly in two ways. \emph{Exact matching} gives the engine and trainer shared
kernels or a single implementation \citep{he2025nondeterminism,vllm2025bitwise,miles2026trueonpolicy}. It must be
rebuilt for each model and GPU type and cannot remove mismatch from quantization with a BF16 trainer or
asynchrony. \emph{Per-token correction} keeps the mismatch and adjusts each sampled token's update:
truncated importance sampling reweights it by a capped probability ratio \citep{yao2025offpolicy,yao2025rollout},
token masks drop it \citep{ling2025ring1t}, and score centering removes the \revorange{drift} \citep{marek2026score}. These
corrections act on sampled tokens, not on which prefixes the engine reaches; for long responses, they
cannot bound the error \citep{li2025trust}.

\noindent
\textbf{Motivation: mismatch budgets in OPD.}
OPD pipelines use these corrections as well
\citep{xiaomi2026mimo,yang2026nemotroncascade2,nvidia2026nemotron3ultra},
but mismatch may affect OPD differently because its update differs from RL's. As shown above, matching the teacher
remains a stationary point whichever engine generates the responses. Mismatch can still change
the path to this point and degrade training quality. This raises three questions:
\textbf{(i)}~How much mismatch can OPD tolerate across different rollout engines and student models?
\textbf{(ii)}~Can we identify safe rollout configurations before training and use them to reduce rollout costs?
\textbf{(iii)}~How does mismatch affect OPD's updates, and can per-token corrections mitigate its effects?

We address these questions by defining a common mismatch measure, an empirical \emph{mismatch budget}, and a
lightweight admission check (\S\ref{sec:overview}). Using controlled rollout configurations and evaluation criteria
(\S\ref{sec:setup}), we measure tolerance, test the admission check, \revorange{quantify the benefits of cheaper rollouts,
and compare OPD with RL (\S\ref{sec:eval}). Finally, we analyze} update bias and the limits of per-token corrections
(\S\ref{sec:discussion}).

%% file: sections/s03-overview.tex
\begin{figure}[t]
\centering
\includegraphics[width=\columnwidth]{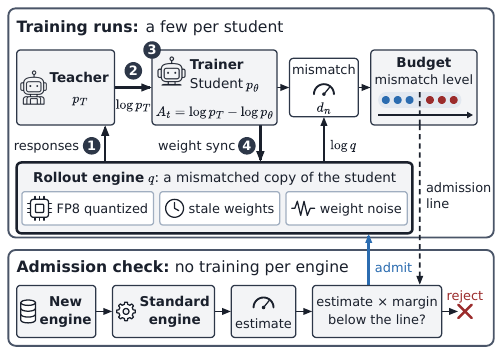}
\caption{\textbf{OPD workflow and mismatch budgets.}
Top: \stepnum{1} the rollout engine generates responses;
\stepnum{2} the teacher scores sampled tokens;
\stepnum{3} the trainer computes advantages and updates the student; and
\stepnum{4} updated weights are synchronized to the engine.
Training runs locate the mismatch budget (\S\ref{sec:budget}); each dot represents a passing (blue) or failing (red) run.
Bottom: the admission check estimates a new engine's mismatch before training
(\S\ref{sec:admission-overview}).\looseness=-1}
\label{fig:overview}\label{fig:diag-system}
\vspace{-10pt}%
\end{figure}
\section{Quantifying Mismatch in OPD}
\label{sec:overview}\label{sec:method}

Cheaper rollouts benefit OPD only if the mismatch they introduce does not degrade training. To study this trade-off,
we measure mismatch on a common scale, determine how much a student tolerates, and estimate the mismatch of new
rollout engines before training. %

\subsection{Measuring Training--Inference Mismatch}
\label{sec:measure}

Let $q$ denote the rollout engine's next-token distribution and $p_\theta$ the current student distribution used by the trainer. As shown in Figure~\ref{fig:overview} (top), we measure mismatch between these distributions
on the prefixes visited by the engine. They may differ because of inference numerics, quantization, or stale weights. The trainer recomputes the student's log-probabilities and the advantage in Eq.~\eqref{eq:bg-adv}; therefore, the engine changes which tokens and prefixes are used for training, not the
advantage assigned to a given token at a given prefix. 

At training step $n$, we measure mismatch over the $N_n$ tokens sampled by the engine:
\begin{equation}\label{eq:dose}
  d_n=\frac{1}{N_n}\sum_t\bigl[\log q(y_t\mid y_{<t})-\log p_\theta(y_t\mid y_{<t})\bigr].
\end{equation}
This estimates the expected per-token $\mathrm{KL}(q\,\|\,p_\theta)$ over the prefixes visited by the engine,
in \emph{nats per token}. An expected value of zero means the two distributions agree at those prefixes. When
rollouts use stale weights, $q$ is the older model that generated them. Thus, the same measure covers numerical
differences, quantization, and staleness.

For a rollout configuration $c$, we summarize its training run by the \emph{mismatch level}
$D(c)$, the average of $d_n$ over its training steps \revorange{(steps 1--200 unless a window is stated)}. In our
setup, the \emph{standard engine} (unmodified vLLM in BF16) has a mismatch of 0.0008~nats/token. This common scale
lets us compare engines whose configurations are otherwise expressed in different terms, such as bit width and
number of stale model versions.

\subsection{Defining the Mismatch Budget}
\label{sec:budget-method}

We call the mismatch that OPD can tolerate without a meaningful loss in training quality its
\emph{mismatch budget}. To characterize it, we train the same student with different rollout
configurations and compare their outcomes against a \emph{standard run} using the standard engine.

For a held-out problem $x$, let $K_x(c)$ denote the student's per-token KL to the teacher
at the evaluation checkpoint after training with configuration $c$, and let $K_x(0)$ denote
the corresponding value for the standard run. All students are evaluated using the standard
engine, rather than their training rollout engine. We report the mean held-out KL as
$K(c)=|\mathcal H|^{-1}\sum_{x\in\mathcal H}K_x(c)$ over the held-out set $\mathcal H$.
To judge relative degradation, we \revorange{divide the mean per-problem change in held-out KL by the standard run's mean}:
\begin{equation}\label{eq:quality-degradation}
  \Delta(c)=\frac{\sum_{x\in\mathcal H}\bigl(K_x(c)-K_x(0)\bigr)}{\sum_{x\in\mathcal H}K_x(0)}
  =\frac{K(c)}{K(0)}-1.
\end{equation}
A positive $\Delta(c)$ indicates worse agreement with the teacher. We classify a run as
failing when $\Delta(c)$ reaches a predefined quality threshold $\tau$ and a paired
statistical test detects a difference; \S\ref{sec:setup-eval} specifies the values and test.

Let $\mathcal P_\tau$ and $\mathcal F_\tau$ be the tested \revorange{runs} that pass and
fail this rule. When the passing and failing levels are ordered, we characterize the
observed transition by an \emph{empirical budget bracket}:
\begin{equation}\label{eq:mismatch-budget}
  B_{\mathrm{emp}}(\tau)=
  \left[\max_{c\in\mathcal P_\tau} D(c),\;
        \min_{c\in\mathcal F_\tau} D(c)\right].
\end{equation}
Figure~\ref{fig:overview} (top right) illustrates this bracket: each dot represents a training run that passes
or fails the quality rule. The bracket locates the transition in the tested configurations; it does not guarantee
that every engine below its upper bound will preserve quality. Its value depends on the
student, training setup, and quality threshold.

\subsection{Estimating Mismatch Before Training}
\label{sec:admission-overview}

Locating a budget requires full training runs, but repeating them for every new rollout
configuration is expensive. We therefore use a lightweight \emph{admission check}
(Figure~\ref{fig:overview}, bottom) to screen new candidates after a student's budget has been characterized. A candidate engine
$c$ generates a small set of responses, and the standard BF16 engine re-scores the
\emph{same tokens} using the same student weights. For the $M$ sampled tokens, we compute
\begin{equation}\label{eq:offline-mismatch}
  \widehat D(c)=\frac{1}{M}\sum_t
  \left[\log q_c(y_t\mid y_{<t})-
        \log q_0(y_t\mid y_{<t})\right],
\end{equation}
where $q_c$ and $q_0$ are the candidate and standard engines' token distributions.
This is an offline estimate of their per-token difference on the candidate's visited
prefixes. It is only a proxy for $D(c)$: the standard engine is not numerically identical
to the trainer, and mismatch may change as the student trains.

We set an admission line $L$ using known passing runs and apply a safety margin
$\alpha>1$ to the offline estimate. The admission rule is
\begin{equation}\label{eq:admission-rule}
  \operatorname{admit}(c)
    =\mathbf{1}\!\left[\alpha\widehat D(c)<L\right].
\end{equation}
An admitted configuration is \emph{predicted} to preserve training quality, not proven
safe. The check requires neither teacher scoring nor a full training run for that
candidate. Its student-specific budget and admission line still require initial
calibration, and we monitor $d_n$ during training, particularly with stale rollouts.
\S\ref{sec:setup-eval} gives the check's experimental settings, and
\S\ref{sec:shortcuts} evaluates its predictions and limitations.

%% file: sections/s04-setup.tex
\section{Experimental Setup}
\label{sec:setup}

Building on the mismatch measure, empirical budget,
and admission check introduced in \S\ref{sec:overview},
we detail the experimental setup used to evaluate them.
We describe the student--teacher pairs, rollout
configurations spanning quantization, stale weights,
and controlled noise, and the evaluation protocols.
We report the results on mismatch tolerance,
admission accuracy, and rollout efficiency in
\S\ref{sec:eval}.

\subsection{Training Setup}\label{sec:setup-train}

\textbf{Models.}
We consider three student models from two model
families, with sizes ranging from 1.5B to 7B
parameters. The students are DeepSeek-R1-Distill-Qwen-7B
and DeepSeek-R1-Distill-Qwen-1.5B
\citep{guo2025deepseekr1}, and Llama-3.2-3B-Instruct
\citep{grattafiori2024llama3}.
Their respective teachers are Skywork-OR1-Math-7B
\citep{he2025skywork}, JustRL-DeepSeek-1.5B
\citep{he2025justrl}, and GT-Llama-3.2-3B-Instruct-MATH.

\textbf{Testbed.}
Our testbed consists of two nodes, each equipped
with four NVIDIA GH200 GPUs. In synchronous runs,
one node hosts the teacher, while the other hosts
the trainer and four rollout replicas. Asynchronous
runs use the same eight GPUs with different
placements (App.~\ref{app:staleness}).

\textbf{Implementation.}
We use vLLM \citep{kwon2023pagedattention} as the
\emph{rollout engine} and verl \citep{sheng2025hybridflow}
for OPD training.
Unless stated otherwise, we use neither importance
correction nor score centering; both are evaluated
in \S\ref{sec:discussion}. 

\textbf{Training recipe.} Each step samples 8 responses per problem for 8 DAPO-Math-17k problems
\citep{yu2025dapo} at temperature 1, with up to 16{,}384 tokens per response. The trainer takes one AdamW
update per step at learning rate $10^{-6}$. Each budget run lasts 200 steps.

\subsection{Rollout Configurations}\label{sec:setup-samplers}

\revorange{On the 7B student, we} study OPD's tolerance to mismatch introduced by
practical rollout optimizations (quantization and
asynchrony), alone and combined. We also
use weight noise to vary mismatch in a controlled
manner. For quantized and noisy configurations, we
modify only the rollout engine; the trainer, teacher,
and checkpoints remain in BF16.

\textbf{(i)}~\emph{Quantization.}
We test three recipes:
(\textit{a})~NVIDIA Model Optimizer FP8
\citep{nvidia2026modelopt}, which quantizes weights,
activations, and KV cache;
(\textit{b})~NVFP4 \citep{alvarez2025nvfp4}, which
uses 4-bit weights and activations with FP8 KV cache;
and (\textit{c})~vLLM's FP8 KV cache, which leaves
weights and activations in BF16.
We test each recipe across all layers and with a
variant retaining the first layer's KV cache in BF16.
For Model Optimizer FP8, this variant instead retains
the entire first and last layers in BF16.
Because NVFP4 requires Blackwell GPUs, we emulate
its quantization in BF16 and evaluate quality only,
not speed (App.~\ref{app:quant}).

\textbf{(ii)}~\emph{Stale weights.}
We vary weight staleness by allowing the rollout
engine to use weights up to
$k\in\{1,2,4,8\}$ model versions old, using verl's
fully asynchronous mode \citep{verl2026fullyasync}.
We also test colocated generation and training,
where rollouts are at most two versions old
(App.~\ref{app:staleness}). \looseness=-1

\textbf{(iii)}~\emph{Weight noise.}
After each weight synchronization, we add a fixed
Gaussian offset to the rollout engine's weights,
scaled by $\sigma\in\{0.0665,0.15,0.234,0.325\}$
times each projection matrix's standard deviation.
This provides controlled perturbations along a fixed
weight-space direction.

\textbf{(iv)}~\emph{Combined configurations.}
We combine Model Optimizer FP8, retaining the first
and last layers in BF16, with rollout weights up to
four or eight versions old.

\subsection{Evaluation and Admission Protocol}
\label{sec:setup-eval}\label{sec:rule}

We specify how we evaluate training quality, classify rollout configurations for the empirical mismatch budget in
\S\ref{sec:budget-method}, and calibrate the admission check in \S\ref{sec:admission-overview}.

\textbf{Held-out KL evaluation.}
We evaluate $K_x(c)$ and $\Delta(c)$ on 530 held-out math problems from AIME24
\citep{maa2024aime} and MATH-500 \citep{hendrycks2021math,lightman2024verify}. At each evaluation checkpoint,
responses are generated by the standard BF16 engine; the student and teacher score the same tokens.
We compare each configuration with the standard run at the same training step.
As defined in Eq.~\eqref{eq:quality-degradation}, $\Delta(c)$ \revorange{is the relative change in mean held-out KL; the
bootstrap interval resamples problems.}

\textbf{Primary quality rule.}
We set the \revorange{quality} threshold to $\tau=0.20$, well above the observed run-to-run variation of at most
1.5\% in final held-out KL (\S\ref{sec:budget}). For scale, standard training on the 7B student reduces
mean held-out KL from 0.0965 to 0.0481; a 20\% increase in the final value would undo about one fifth
of this gain. A configuration fails if $\Delta(c)\ge\tau$ and the 95\% problem-paired bootstrap interval
for its KL gap from the standard run excludes zero. The rule was fixed before the runs it evaluates
(App.~\ref{app:prereg}).

\textbf{Secondary accuracy rule.}
We also evaluate mean accuracy on AIME24, AIME25, and the non-overlapping subset of AMC23, using 16 samples
per problem at step 200. Accuracy is secondary because OPD optimizes teacher agreement rather than this
accuracy measure directly. A run fails this rule if accuracy drops by at least five percentage points
and the 95\% interval for the difference excludes zero. Typical intervals are about $\pm2$ points wide.
Only the primary \revorange{quality} rule determines $B_{\mathrm{emp}}(\tau)$.

\textbf{Admission-check settings.}
\label{sec:setup-admission}\label{sec:admission-method}
We use three settings for the admission check
in \S\ref{sec:admission-overview}.

\textbf{(i)}~\emph{Sample size.}
Each candidate generates two responses for 16 math
problems, re-scored by the standard BF16 engine.
Across six configurations, 95\% interval half-widths
are within 30\% of the offline estimates.

\textbf{(ii)}~\emph{Safety margin.}
A 100-step pilot on the 7B student found in-training
mismatch at most 1.47 times the offline estimate
(App.~\ref{app:pilot}). We thus initially set
$\alpha=1.5$.

\textbf{(iii)}~\emph{Admission line.}
\revorange{Two pilot runs of one quantized engine passed the primary
quality rule at step 100 at about 0.41 nats/token}, so we initially set
$L=0.41$ nats/token. We admit a candidate if
$\widehat D(c)<L/\alpha\approx0.27$ nats/token.
This is a screening rule, not a quality guarantee
(\S\ref{sec:shortcuts}). Each new student requires
budget calibration, and we monitor mismatch during
training.

%% file: sections/s05-evaluation.tex
\begin{figure*}
\centering
\includegraphics[width=\textwidth]{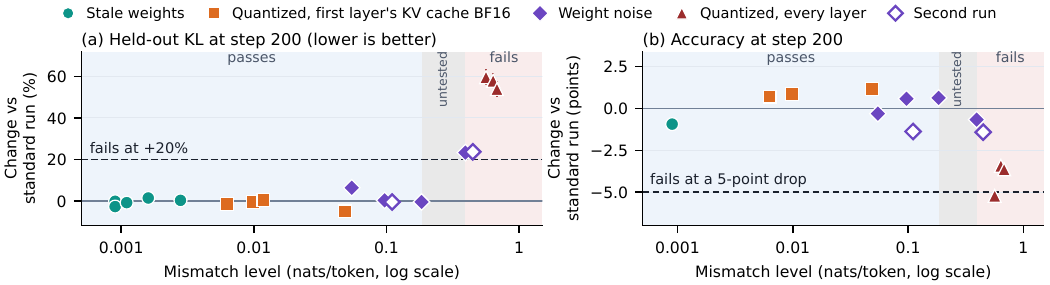}
\caption{\textbf{On the 7B student, runs pass up to 0.184~nats/token and fail at 0.39 and above.}
One point per run; hollow markers are second runs, nudged right. Bars in (a) show 95\% intervals; shading
marks the held-out KL \revorange{verdicts}. Panel (b) includes only runs evaluated for accuracy. Values: App.~Table~\ref{tab:budget}.}
\label{fig:budget}
\end{figure*}

\section{Characterizing Mismatch Budgets}
\label{sec:eval}

Building on the methodology in \S\ref{sec:overview} and setup in \S\ref{sec:setup}, we investigate
\revorange{questions (i) and (ii) of \S\ref{sec:background} and compare OPD with RL:} \textit{(i) Mismatch tolerance:} How much mismatch can OPD tolerate (\S\ref{sec:budget}--\S\ref{sec:students})?
\textit{(ii) Admission and efficiency:} Can we screen configurations before full training, and what
benefits do they provide (\S\ref{sec:shortcuts}--\S\ref{sec:speed})?
\textit{OPD versus RL:} Do configurations tolerated by OPD also preserve RL training quality
(\S\ref{sec:rl-contrast})?

\subsection{Quantifying OPD's Mismatch Budget}
\label{sec:budget}

We measure the 7B student's tolerance by training with quantized, stale-weight, and noisy rollout
engines and comparing their mismatch $D(c)$ and held-out KL with standard BF16 under the primary
quality rule (\S\ref{sec:setup-eval}). Figure~\ref{fig:budget} shows 18 runs \revorange{of 16 configurations, two run twice; repeated runs
differ in final KL} by at most 1.5\% (App.~\ref{app:prereg:onerun}).

\textbf{(1)}~\emph{OPD tolerates substantial mismatch} \revorange{(Figure~\ref{fig:budget}; App.~Table~\ref{tab:budget})}. All twelve configurations with
$D(c)\le0.184$~nats/token, over 200 times the standard engine's $D(0)=0.0008$,
pass with $\Delta(c)$ between $-5.0\%$ and $+6.4\%$. \revorange{Under the same engines, REINFORCE and GRPO break at about 12 and
60 times $D(0)$ (\S\ref{sec:rl-contrast}).}
\revorange{Even the two highest passing levels finish within 0.5\% of the standard run's held-out KL.}

\textbf{(2)}~\emph{Larger mismatch degrades quality} \revorange{(Figure~\ref{fig:budget}; App.~Table~\ref{tab:budget})}. Both noise runs at $D(c)\approx0.39$
fail with $\Delta(c)\approx+23\%$, giving the empirical bracket
$B_{\mathrm{emp}}(0.20)=[0.184,0.39]$~nats/token, not an exact failure threshold.
\revorange{The three every-layer quantized configurations, with the first layer's KV cache also in FP8, reach
$D(c)=0.56$--$0.68$, fail with $\Delta(c)=+54\%$--$+60\%$, and lose 3--5 accuracy points; under every-layer
FP8, rollout answers are almost never correct from the start (Figure~\ref{fig:intro-accuracy-time}).
The bracket depends on the quality rule: at $\tau=0.25$ it is $[0.39,0.56]$ (App.~\ref{app:sensitivity}).}

\textbf{(3)}~\emph{The measure compares different mismatch sources} \revorange{(Figure~\ref{fig:budget})}. Passing and failing levels
are broadly consistent across quantization and weight noise. Staleness of up to eight
versions yields only $D(c)=0.0028$, far below the failure range. Neither bit width nor \revorange{staleness}
alone captures the resulting mismatch.

\textbf{Insight.}
A common mismatch measure helps characterize tolerance across rollout optimizations, but the
observed budget is specific to the student and quality rule.

\subsection{Model-Dependent Mismatch Tolerance}
\label{sec:students}

We examine whether the 7B findings transfer to the 1.5B and 3B students by comparing FP8-induced
mismatch and training quality. We also vary precision across components on the 7B student.
These tests do not establish complete budget brackets for the smaller students.

\textbf{(1)}~\emph{Similar mismatch need not imply similar quality loss} \revorange{(App.~Table~\ref{tab:admission})}. \revorange{Two every-layer FP8 recipes
on the 1.5B student yield $D(c)=0.36$--$0.39$ over steps 1--100, versus
$0.65$--$0.69$ for comparable 7B quantization, yet the 1.5B runs finish with
$\Delta(c)=+420\%$ to $+479\%$ and about 11 points lower accuracy; 7B weight-noise runs at a similar
mismatch of 0.39 have $\Delta(c)\approx+23\%$.} Because both model and mismatch
source differ, this comparison does not isolate model tolerance; it argues against
transferring the 7B budget without calibration. Keeping the 1.5B student's first-layer KV
cache in BF16 avoids the observed degradation ($\Delta(c)=-4.3\%$).

\begin{figure}[t]
\centering
\includegraphics[width=\columnwidth]{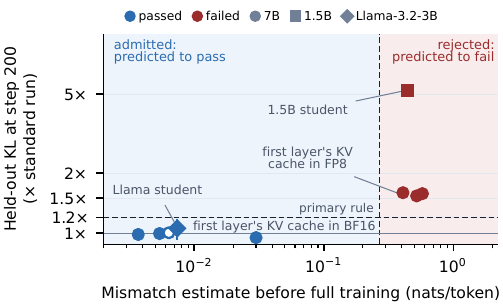}
\caption{\textbf{Mismatch estimates separate the tested passing and failing configurations.}
Points show trained engines, their estimated mismatch, and final held-out KL; error bars show 95\% intervals.
For the 1.5B and Llama points, estimates come from five-step training runs, not offline checks.
Dashed: the pilot admission boundary $L/\alpha\approx0.27$~nats/token. Hollow: an engine with weights up to
eight versions old, shown at its recipe's estimate.}
\label{fig:admission}
\end{figure}

\textbf{(2)}~\emph{FP8 mismatch depends on the model and recipe} \revorange{(Figure~\ref{fig:admission}; App.~Table~\ref{tab:admission})}. On the 7B
student, retaining the first-layer KV cache in BF16 reduces vLLM's FP8-cache \revorange{estimate $\widehat D(c)$} from
0.52 to 0.0037~nats/token. By contrast,
\revorange{every-layer} Model Optimizer FP8 on Llama-3.2-3B yields only $D(c)\approx0.01$ and passes
($\Delta(c)=+5.6\%$). An offline Qwen3-4B check with
the score-centering FP8 recipe \citep{marek2026code} gives $\widehat D(c)=0.005$ \revorange{(App.~\ref{app:quant:offline})}.

\begin{figure}[t]
\centering
\includegraphics[width=\columnwidth]{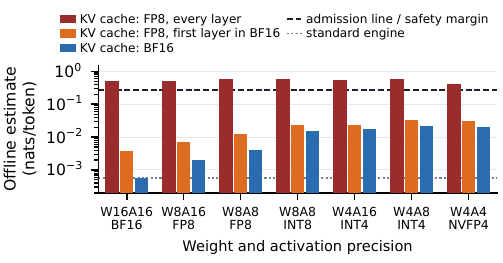}
\caption{\textbf{On the 7B student, an FP8 KV cache dominates mismatch, and one layer carries it.}
W8A16 means 8-bit weights and 16-bit activations. Dashed: \revorange{admission boundary $L/\alpha\approx0.27$; engines below it are admitted}. Recipes: App.~Table~\ref{tab:app-recipes}.}
\label{fig:shortcuts}
\end{figure}

\textbf{(3)}~\emph{Mismatch measurements locate sensitive components} \revorange{(Figure~\ref{fig:shortcuts})}. On the 7B student,
\revorange{an FP8 KV cache dominates the mismatch at every tested weight and activation precision, and
the first layer's cache accounts for 93\%--99\% of it across the seven FP8-cache recipes in Figure~\ref{fig:shortcuts}.}
The DeepSeek students have a key-projection bias and lack query/key normalization
\citep{henry2020qknorm}; on the 7B student, \revorange{the first layer's keys are much larger than in most layers}. A keys-only emulation attributes about 82\% of that layer's effect
to first-layer key rounding (App.~\ref{app:firstlayer}). Llama lacks key-projection bias,
while Qwen3 normalizes queries and keys, consistent with their lower sensitivity.

\textbf{Insight.}
Both the mismatch introduced by a recipe and its quality impact depend on the student;
candidate engines therefore require student-specific measurement.

\subsection{Evaluating the Admission Check}
\label{sec:shortcuts}\label{sec:hazard}

We compare pre-run mismatch estimates $\widehat D(c)$ and admission decisions with full-training
outcomes, then assess the check's cost and pilot-derived parameters (\S\ref{sec:setup-admission}).

\textbf{(1)}~\emph{All eight preselected decisions match the observed outcomes} \revorange{(Figure~\ref{fig:admission}; App.~Table~\ref{tab:admission})}.
\revorange{Of the nine trained configurations in Figure~\ref{fig:admission}, all eight with decisions fixed before full training,
five from the offline check, match the primary quality rule.} On the 7B student, retaining first-layer
BF16 KV cache passes while quantizing it to FP8 fails; offline estimates fall at
$\widehat D(c)\le0.030$ or $\ge0.408$. The boundary $L/\alpha\approx0.27$ lies between these
groups, so \revorange{the check is untested near the boundary}.

\textbf{(2)}~\emph{Screening is inexpensive after calibration.} Checking six configurations
takes 24 minutes on one GPU, versus \revorange{5.5 to 6 hours} on eight GPUs for one full training run.
The check requires neither trainer nor teacher, but the initial student-specific budget
calibration still requires representative training runs.

\textbf{(3)}~\emph{Pilot-based safety settings can be insufficient} \revorange{(App.~Table~\ref{tab:admission})}. During training, \revorange{the admitted 7B}
engines reach \revorange{1.56--1.68 times their estimates (steps 1--100)}, exceeding the pilot margin $\alpha=1.5$. Noise at $D(c)=0.39$ fails \revorange{at step 200} despite lying below the pilot
admission line $L=0.41$. Retrospectively setting $L=0.184$ would preserve all eight tested
decisions with a more conservative boundary, but does not establish general safety.

\textbf{Insight.}
Offline estimates support low-cost screening, provided the admission line and margin are
calibrated conservatively and mismatch is monitored during training.

\subsection{Efficiency Within the Budget}
\label{sec:speed}

We measure throughput, KV-cache capacity, and time-to-quality on the 7B student using
quality-preserving rollout configurations. Throughput is optimizer \emph{updates per hour}
relative to synchronous BF16, excluding warm-up and asynchronous drain-only steps
(App.~\ref{app:prereg:speed}). We also benchmark a separate memory-bound batch.

\begin{figure}[t]
\centering
\includegraphics[width=\columnwidth]{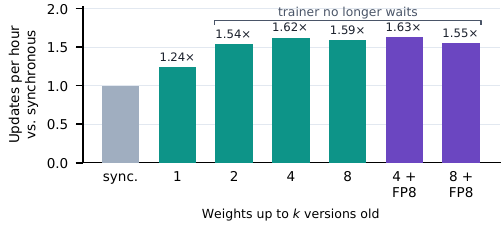}
\caption{\textbf{Asynchronous rollouts improve throughput until the trainer no longer waits.} One run per bar
on the 7B student. Rollout weights can be up to $k$ versions older than the trainer's; purple bars additionally
use Model Optimizer's FP8.}
\label{fig:speed-async}
\end{figure}

\begin{figure}[t]
\centering
\includegraphics[width=\columnwidth]{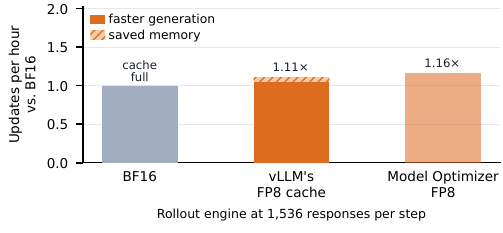}
\caption{\textbf{At a memory-bound rollout batch, FP8 benefits from faster generation and saved KV-cache memory.}
The 7B student with 1{,}536 responses per step, 24 times the training batch. The lighter Model Optimizer FP8
bar is not decomposed. Values: App.~Table~\ref{tab:speed-samplers}.}
\label{fig:speed-fp8}
\end{figure}

\textbf{(1)}~\emph{Asynchrony helps until the trainer becomes the bottleneck} \revorange{(Figure~\ref{fig:speed-async}; App.~Table~\ref{tab:speed-async})}.
\revorange{Throughput rises with staleness $k$ and levels off at $1.62\times$ and $1.59\times$ for $k=4$ and 8;
all runs pass and finish within 3\% of the standard held-out KL. In synchronous training, the trainer
waits for rollouts and teacher scores (App.~Figure~\ref{fig:speed-steptime}); from $k=2$ on, this wait is at
most three seconds, and training limits throughput to about 55 updates/hour. FP8 combined
with $k=4,8$ gives $1.63\times$ and $1.55\times$, close to asynchrony alone, because generation is off the
critical path.}

\textbf{(2)}~\emph{FP8 weights improve speed; FP8 KV cache mainly adds capacity} \revorange{(Figure~\ref{fig:speed-fp8}; App.~Tables~\ref{tab:speed-samplers} and~\ref{tab:speed-64})}.
At 64 responses/step, Model Optimizer FP8 with BF16 first/last layers \revorange{speeds up generation} and raises
updates/hour by $1.27\times$; \revorange{vLLM's FP8 cache} alone gives $1.00\times$
(App.~Table~\ref{tab:speed-64}). FP8 holds at least $1.93\times$ as many KV-cache tokens\revorange{ on the 7B student,} but BF16 does not exhaust its cache at this batch size.
At 1{,}536 responses/step, \revorange{BF16 runs out of cache and must evict and recompute running responses, while
FP8 does not. vLLM's FP8 cache} and Model Optimizer FP8 then give
$1.11\times$ and $1.16\times$ throughput, respectively.
The latter gain is smaller than at the training batch because generation and teacher
scoring take only 42\% of each step (App.~\ref{app:speedbench}). 

\textbf{(3)}~\emph{Cheaper rollouts reach the same quality sooner} \revorange{(App.~Table~\ref{tab:speed-ttq})}. We measure wall-clock
time from the first training step to the synchronous standard run's step-200 held-out KL,
including generation, teacher scoring, and checkpoint saves. \revorange{Model Optimizer FP8 with $k=8$
reaches this target $1.62\times$ sooner, with final accuracy 60.5\% versus 61.5\%.}
A second preselected target, 90\% of the standard run's KL improvement, gives the same
speedup; with FP8 and $k=4$, it is reached $1.69\times$ sooner and final
KL stays within 0.8\% of the standard run.

\textbf{Insight.}
\revorange{Asynchrony removes rollout waiting, while FP8 benefits depend on whether generation speed
or KV-cache capacity limits the workload. All these 200-step runs pass the primary rule with
$D(c)\le0.013$~nats/token.}

\begin{figure*}[!t]
\centering
\includegraphics[width=\textwidth]{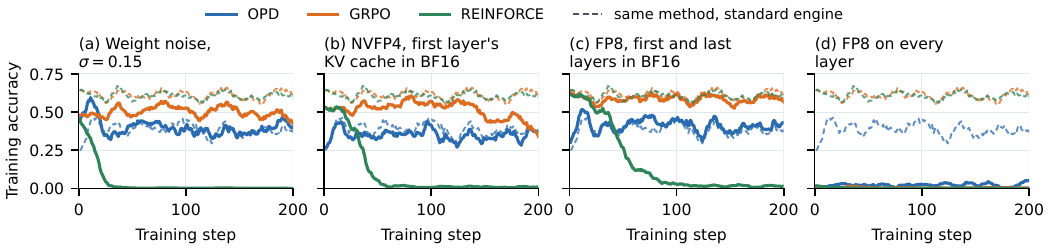}
\caption{\textbf{REINFORCE broke under all three rollout configurations that OPD tolerates, and GRPO under two by step 200.}
Training accuracy (for GRPO and REINFORCE, mean 0/1 reward), smoothed over 11 steps. Each method uses different
prompts, so compare it only with its own dashed standard-engine baseline.}
\label{fig:rl-contrast}
\end{figure*}

\subsection{RL Breaks Where OPD Holds}
\label{sec:rl-contrast}

We test whether configurations tolerated by OPD also preserve RL training quality by
comparing GRPO and REINFORCE with their respective standard-engine baselines (Figure~\ref{fig:rl-contrast}); prior work reports RL mismatch
sensitivity \citep{zhong2026vexact,liu2025speedkills}.

\paragraph{Setup.}
Both RL methods train the 7B student with a 0/1 reward for a correct boxed answer,
without a teacher. Following a scaled-down JustRL recipe \citep{he2025justrl}, each step
uses 16 problems and eight responses per problem, learning rate $10^{-6}$, and no KL penalty
or importance correction. GRPO \citep{shao2024deepseekmath} normalizes within each group
and takes two updates per step; REINFORCE \citep{williams1992reinforce,ahmadian2024back}
uses raw reward. Each method runs 200 steps with four OPD rollout configurations and its
own standard engine. A run \emph{breaks} if held-out DAPO accuracy on 1{,}000 problems
drops at least five points, or if entropy or response length collapses with held-out
confirmation. These criteria were fixed in advance. Neither standard RL run improves
held-out accuracy measurably by step 100, so we assess failure rather than learning
gains (App.~\ref{app:prereg:rl}).

\textbf{(1)}~\emph{REINFORCE breaks under all three OPD-passing configurations} \revorange{(Figure~\ref{fig:rl-contrast})}.
\revorange{With weight noise at 0.097 ($\sigma=0.15$), its reward reaches zero and
held-out accuracy falls 21 points by step 200. With NVFP4 and first-layer BF16 KV cache, responses
hit the length limit and accuracy falls 59 points. With FP8 keeping
the first and last layers in BF16, reward nearly vanishes and accuracy falls
63 points.}

\textbf{(2)}~\emph{GRPO breaks under two of the three} \revorange{(Figure~\ref{fig:rl-contrast})}. Noise and NVFP4 break it, but
\revorange{FP8 with BF16 first and last layers passes. With NVFP4, entropy collapses
and accuracy falls 15 points; with noise, more held-out answers hit the length limit.
Every-layer FP8 breaks all methods, including OPD.}

\textbf{(3)}~\emph{Initial mismatch and early weight changes do not explain the contrast.}
RL runs start at 0.88--1.02 times the matching OPD runs' mismatch and move
weights only 0.43--0.90 times as far \revorange{as OPD's standard run} in the first 100 steps.

\textbf{(4)}~\emph{Score centering prevents the tested RL failures} \revorange{(App.~\ref{app:rl-sc})}. Under one engine per
method, score centering \citep{marek2026score} raises REINFORCE's held-out accuracy by
61 points at step 100 and GRPO's by 16.5 points at step 200. A control that computes
but does not apply centering still breaks.

\textbf{Insight.}
The tested RL failures do not predict OPD's tolerance. We examine this contrast and
the limits of per-token corrections in \S\ref{sec:discussion}.

%% file: sections/n06-discussion.tex
\section{Discussion}
\label{sec:discussion}\label{sec:mechanism}\label{sec:corrections}

OPD's target explains part of its tolerance\revorange{; the drift's sign does not. Per-token} corrections do not help OPD, \revorange{and} under weight noise part of the error comes from the prefixes (partial
responses) that the engine reaches.

\textbf{(1)}~\emph{The error vanishes at the teacher.} At one prefix, let $\delta=q-p_\theta$ be the engine's \revorange{deviation}
and $A=\log p_T-\log p_\theta$ the teacher--student log ratio over the vocabulary. In the gradient with respect to the
logits $z$, mismatch adds
\begin{equation}\label{eq:mech-error}
\begin{gathered}
  g^{z}_{\mathrm{bias}}=\delta\odot A-p_\theta\,(\delta^{\top}A),\\
  \bigl\|g^{z}_{\mathrm{bias}}\bigr\|_1\le 2\,\|\delta\|_1\,\|A\|_\infty.
\end{gathered}
\end{equation}
It vanishes as the student approaches the teacher, whatever the engine. With a constant reward $R$ in place of $A$, as
in REINFORCE, the error is $R\,\delta$, which need not shrink. With the engine fixed, the expected OPD gradient is
exactly that of $\tfrac12\,\mathbb{E}_{q}[(\log p_\theta-\log p_T)^2]$ (App.~\ref{app:anchor}), so matching the
teacher stays a stationary point whatever the engine. This needs the advantage computed from the trainer's
log-probabilities, not the engine's.

\textbf{(2)}~\emph{During training, the error shrinks, but the update shrinks faster.} \revorange{At three weight-noise levels on fixed prefixes, the error of
Eq.~\eqref{eq:mech-error} at step 200 is 0.11 to 0.13 of its size at the untrained student. The update itself shrinks faster:
at the middle level, the error-to-update ratio is} 0.93 at the untrained student and 1.88 at step 200 (App.~\ref{app:probe}). So
mismatch matters more, not less, late in training.

\textbf{(3)}~\emph{The drift's sign does not protect OPD.} Under weight noise, the student moves away from the engine,
as OPD's negative drift coefficient \revorange{(the mean advantage under the engine)} predicts (\revorange{App.~\ref{app:prereg:rules}}). \revorange{Adding} a constant
$b$ to every advantage \revorange{keeps the expected on-policy update but shifts the coefficient by $b$}. With $b=+0.15$ and $+0.5$ \rev{under noise at mismatch} 0.097, the coefficient turns positive, as in REINFORCE, and
the student moves \emph{toward} the engine, \rev{but} held-out KL ends 6.5\% and 6.9\% below the standard run's \revorange{(App.~Table~\ref{tab:mech-tests})}.

\textbf{(4)}~\emph{\revorange{Neither tested correction helps OPD}.} Score centering \citep{marek2026score} removes the drift,
but \rev{under noise at mismatch} 0.055 and 0.097 it leaves held-out KL 2.3\% to 3.7\% \revorange{above uncentered OPD's}. At 0.097,
truncated importance sampling raises held-out KL by 2.9\% and 7.0\% at steps 100 and 200, and FP8 on every layer still
fails.

\begin{figure}[t]
\centering
\includegraphics[width=\columnwidth]{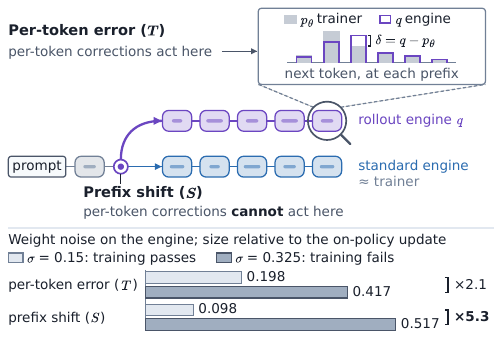}
\caption{\textbf{Two ways mismatch adds error to OPD's update.} \revorange{Boxes are tokens; the rollout engine's response (purple) leaves the standard engine's (blue) at the fork.} \textbf{Bottom}: \revorange{the size of $T$ and $S$ at the untrained student, to scale.}}
\label{fig:prefix-shift}\label{fig:diag-prefix-shift}
\end{figure}

\textbf{(5)}~\emph{Part of the error comes from the prefixes the engine reaches.} We split the update's error into two
parts (Figure~\ref{fig:prefix-shift}; App.~\ref{app:shift}): $T$, the \emph{per-token error}, from the engine's different next-token
probabilities at the prefixes it visits, and $S$, the \emph{prefix shift}, from visiting different prefixes than the
standard engine. Both parts grow from \revorange{a passing noise level (0.097, $\sigma=0.15$) to the failing one (0.39, $\sigma=0.325$)}. At the untrained student, $S$ grows more: 5.3 times against 2.1 in our main split, and more in every
way of splitting. Per-token corrections act only on $T$. Only reweighting whole responses reaches $S$, and by our
estimate that needs about $10^{27}$ responses per effective sample even at our smallest noise
(App.~\ref{app:corrections}), so keeping mismatch inside the budget remains the practical remedy.

\textbf{(6)}~\emph{Offline OPD sits inside the budget.} Semi-OPD and Lightning OPD \citep{zhao2026semiopd,wu2026lightning}
draw rollouts once from the untrained student and rescore them with the current student, \revorange{in effect using a rollout engine with fixed weights}. Neither paper reports a KL mismatch. On \revorange{7B standard-run checkpoints} it
is 0.082~nats/token at steps 100 and 200, below the \revorange{bracket's} lower end of 0.184 (App.~\ref{app:frozen}).

%% file: sections/n07-related.tex
\section{Related Work}
\label{sec:related}

\textbf{Cheaper rollouts.}
Quantized rollouts reduce memory costs but introduce training--inference mismatch. Prior RL work explores update correction, shared low-precision execution, and quantization-aware approaches \citep{liu2025flashrl,qiu2026fp8rl,li2026qurl,huang2025qerl,xi2026jetrl,gu2026qarl}. KV-cache quantizers preserve sensitive tokens or outliers \citep{liu2024kivi,hooper2024kvquant}, while asynchronous RL overlaps generation and learning at the cost of stale weights \citep{fu2025areal,verl2026rlinsight,hu2026flashreinforce}. \citet{zheng2025stabilizing} factor one importance weight into mismatch and staleness. We compare how quantization, staleness, and controlled weight noise affect OPD training quality.

\textbf{Removing mismatch.}
Exact-matching approaches align rollout and training execution through shared or deterministic kernels and unified execution \citep{he2025nondeterminism,vllm2025bitwise,miles2026trueonpolicy,zhang2025tbik,jiang2026isoexec,vllm2026consistency}. Routing replay similarly aligns expert choices in mixture-of-experts models \citep{ma2025r3}. Such approaches address execution-induced mismatch, whereas quantizing only the rollout engine or using stale weights introduces mismatch by design. Rather than requiring exact agreement, this study measures how much mismatch OPD can tolerate.

\textbf{Correcting the update in RL.}
RL systems mitigate mismatch through importance weighting \citep{yao2025offpolicy,yao2025rollout,yu2026cis,zhou2026ais}, token or response masking \citep{liu2025speedkills,luo2026p1vl,ling2025ring1t,deepseekai2025v32}, and score centering, which removes mismatch-induced drift \citep{marek2026score} (App.~\ref{app:prereg:rules}). Other approaches penalize policy divergence \citep{tinker2026stablereinforce}, while prior work identifies limits of per-token corrections on long responses \citep{li2025trust}. Production OPD pipelines also adopt RL-style corrections \citep{xiaomi2026mimo,yang2026nemotroncascade2,nvidia2026nemotron3ultra}. In tested OPD runs, neither truncated importance sampling nor score centering helps; under weight noise, per-token corrections cannot address prefix shifts (Section~\ref{sec:discussion}).

\textbf{Distillation on stale or fixed rollouts.}
Near-Policy, AsyncOPD, and $f$-OPD study OPD with stale rollouts \citep{rang2026nearpolicy,kang2026asyncopd,chen2026fopd}: AsyncOPD examines KL objectives, while $f$-OPD regulates stale-sample influence. Semi-OPD and Lightning OPD instead reuse rollouts sampled once from the initial student \citep{zhao2026semiopd,wu2026lightning}. Low-bit OPD distills a quantized student \citep{chen2026train}, whereas here the student remains in BF16 and only the rollout engine is quantized. Complementing these methods, this study measures OPD's tolerance across mismatch sources and uses a calibrated budget to screen rollout configurations before training.

%% file: sections/n08-conclusion.tex
\section{Conclusion}
\label{sec:conclusion}

This paper examined the trade-off between rollout efficiency and training quality under training--inference mismatch in OPD. OPD tolerates mismatch that destabilizes RL, but safe ranges depend on the model and cannot be inferred from bit width or staleness alone. \revorange{On the 7B student, OPD keeps its training quality up to more than 200 times a standard engine's mismatch, while two RL methods break at about 12 and 60 times.} Under weight noise, \revorange{part of OPD's error comes from the partial responses the engine reaches, a prefix shift that per-token corrections cannot fix.} Calibrated budgets and an admission check help select FP8 and asynchronous configurations that preserve quality, increasing KV-cache capacity and reaching the same held-out KL target 1.62 times sooner than synchronous BF16 training. Overall, mismatch budgets offer a practical basis for balancing rollout efficiency and training quality in OPD. \revorange{In practice, a few training runs calibrate a student's budget; the admission check then screens new engines in minutes, and logging mismatch throughout training catches any later drift beyond the offline estimate.}

%% file: sections/a-about.tex
\appendix
\section{About This Appendix}
\label{app:about}
This appendix accompanies the main paper, on how much training--inference mismatch on-policy
distillation (OPD) tolerates and what speed that buys. Sections~1--\ref{sec:conclusion},
Figures~1--\ref{fig:prefix-shift} and Equations~(1)--(\ref{eq:mech-error}) are in the main paper; citations refer to
its reference list.

\paragraph{Terms.}
\begin{itemize}
\setlength{\itemsep}{1pt}\setlength{\parskip}{0pt}
\item \emph{Students.} Our 7B student, DeepSeek-R1-Distill-Qwen-7B, has Skywork-OR1-Math-7B as its teacher; the
  smaller students are DeepSeek-R1-Distill-Qwen-1.5B and Llama-3.2-3B-Instruct (Section~\ref{sec:setup}).
\item \emph{Rollout engine, trainer.} The engine (vLLM) generates the training responses from a copy
  of the student; the trainer updates the student. Our engines differ from the trainer by quantization, stale weights
  (asynchronous rollouts) or weight noise.
\item \emph{Standard engine, standard run.} The standard engine is vLLM in BF16 with the trainer's current
  weights; the standard run trains with it.
\item \emph{Mismatch.} Per-token KL from the engine's next-token distribution to the trainer's, on the
  engine's tokens, in nats/token (Eq.~\eqref{eq:dose}); a run's level is its mean over steps (standard engine:
  0.0008).
\item \emph{Held-out KL.} The student's per-token KL to the teacher on 530 held-out math problems, with
  standard-engine responses, in nats/token; verdicts use its change relative to the standard run at the same step.
\item \emph{Primary rule, budget.} A run fails if its held-out KL is 20\% or more above the standard run's
  and the 95\% interval excludes zero (App.~\ref{app:prereg:rules}). The budget lies between the highest passing and
  lowest failing levels: 0.184 to 0.39~nats/token on our 7B student.
\item \emph{Admission check.} An offline mismatch estimate before training; an engine is admitted if 1.5
  times the estimate is below 0.41~nats/token, a line set from the pilot that later proved too high
  (Section~\ref{sec:admission-method}).
\item \emph{Draft head.} The 7B pair's training rollouts use an EAGLE-3 draft head for speculative decoding,
  which leaves the engine's distribution unchanged (App.~\ref{app:repro}).
\end{itemize}

\paragraph{Contents.}
App.~\ref{app:prereg}: pre-registered rules; \ref{app:recipes}: rollout engines; \ref{app:probe}: gradient
probe; \ref{app:speedbench}: speed benchmark; \ref{app:pilot}: the 100-step pilot; \ref{app:repro}: reproducibility;
\ref{app:tables}: result tables; \ref{app:limitations}: limitations.

%% file: sections/a-prereg.tex
\section{Pre-registration}
\label{app:prereg}

This appendix gives the rules behind the main text's verdicts. Each experiment entered a dated, append-only
plan before its jobs ran, except the 30-step check without the draft head (Table~\ref{tab:speed-64}).

\subsection{Rules for the OPD runs}
\label{app:prereg:rules}

These rules judge every OPD run; their verdicts place the budget of Section~\ref{sec:budget}.

\paragraph{Reference and mismatch.}
Each OPD run is compared with a standard run sharing its seed, data order and recipe. At step 200 this is
the 200-step standard run; at step 100, the mean of three earlier 100-step standard runs. Where a plan fixed the step-100
reading as the verdict before training was extended to 200 steps, both readings are reported. The two references differ by
0.04\% in held-out KL at step 100. The staleness runs (App.~\ref{app:staleness}) use the 200-step run at both steps.
A run's mismatch $D$ is verl's per-step estimate of $\mathrm{KL}(q\,\|\,p)$ (Eq.~\eqref{eq:dose}), averaged over the
steps before the checkpoint being read.

\paragraph{Primary rule: held-out KL.}
Each checkpoint samples responses with a standard engine (8 per AIME24 problem, 1 per MATH-500 problem),
which are teacher-forced through the checkpoint and the teacher. $\mathrm{KL}_x$ is the mean of $\log p_\theta(y_t)-\log
p_T(y_t)$ over problem $x$'s response tokens, with AIME24's 8 samples pooled. With
$\mathrm{rel}=\overline{(\mathrm{KL}^{\mathrm{run}}_x-\mathrm{KL}^{\mathrm{ref}}_x)}\,/\,\overline{\mathrm{KL}^{\mathrm{ref}}_x}$
over the 530 problems, a run fails if $\mathrm{rel}\ge+20\%$ and the 95\% percentile interval of the mean difference,
from 10{,}000 bootstrap resamples of problems, excludes zero.

\paragraph{Accuracy composite (secondary).}
The composite averages AIME24, AIME25 (30 problems each) and the 28 of AMC23's 83 problems that a
text-overlap rule does not flag as overlapping the training pool. A run fails if it is at least 5 points below the
reference's with a problem-paired 95\% interval excluding zero.

\paragraph{How the budget depends on the 20\% line.}\label{app:sensitivity}
Any line from 6.4\% to 23.3\% gives the same bracket, [0.184, 0.39] (Table~\ref{tab:sensitivity}). The
accuracy rule is coarser: its intervals are about $\pm2$ points wide (for the runs at 0.39, $-0.66$ [$-2.80$, $+1.50$] and $-1.41$ [$-3.46$, $+0.71$]).

\begin{table}[ht]
\centering\footnotesize
\caption{\textbf{The budget on our 7B student for other held-out KL lines} (step 200, same interval condition; from
Table~\ref{tab:budget}). Fail: runs that fail. Not ordered: a failing level lies below a passing one.}
\label{tab:sensitivity}
\setlength{\tabcolsep}{4pt}
\begin{tabular}{@{}lcp{4.3cm}@{}}
\toprule
Held-out KL line & Fail & Budget bracket (nats/token) \\
\midrule
up to 6.35\% & 6 & not ordered: 0.055 fails, 0.097 and 0.184 pass \\
6.35--23.28\% & 5 & [0.184, 0.39], as in Section~\ref{sec:budget} \\
23.28--23.69\% & 4 & 0.39 splits: one of its two runs fails \\
23.69--53.92\% & 3 & [0.39, 0.56] \\
53.92--59.66\% & 1--2 & not ordered: 0.56 fails, 0.68 passes \\
above 59.66\% & 0 & no run fails \\
\bottomrule
\end{tabular}
\end{table}

\paragraph{Collapse rule.}
The trainer's entropy, the response length or the length-cap rate collapses if its 5-step rolling mean,
minus its mean over the run's first 10 steps, stays outside its band, $[m-R,\,m+R]$, for at least 5 consecutive steps,
with $R$ and $m$ the range and midpoint of the same difference in the reference run.

\paragraph{Drift tests (weight-noise runs).}
These tests ask whether the weights move toward or away from the noisy engine. A checkpoint's displacement at step $s$ along a unit direction $u$, relative to the reference run, is
\begin{equation}
  X_s(u)=\bigl\langle\theta_s^{\mathrm{run}}-\theta_s^{\mathrm{ref}},\,u\bigr\rangle ,
  \label{eq:app-drift}
\end{equation}
with $\hat n$ the direction of the engine's weight offset ($X_s(\hat n)>0$: toward the engine) and
$u_1,\dots,u_{200}$ random directions (App.~\ref{app:noise}). It is \emph{specific} if $|X(\hat n)|$ at step 200
exceeds the 97.5th percentile of $|X(u_k)|$, and has \emph{leveled off} if also specific at step 100 with an increment
to step 200 under half the step-100 value. The first runs at $\sigma=0.0665$, 0.15 and 0.325 read specific, away from
the engine, and leveled off (Section~\ref{sec:mechanism}).

\paragraph{Score centering and the advantage offset.}
Score centering~\citep{marek2026score} replaces the score $\nabla\log p_\theta(y)$ by its difference from
the engine's expected score, which removes the drift term: the mean advantage under the engine, $\bar A_q$, times
the engine's mean score (Section~\ref{sec:discussion}). For OPD, $\bar A_q=\mathrm{KL}(q\,\|\,p_\theta)-\mathrm{KL}(q\,\|\,p_T)$,
negative while the engine is closer to the student than to the teacher; its token means under weight noise are in
Table~\ref{tab:app-probe}.

The advantage offset adds a constant $b$ to every response token's advantage, which shifts the drift
coefficient by $b$ under mismatch (Section~\ref{sec:discussion}). The runs use $b=\pm0.15$ and $b=+0.5$; a positive
offset is predicted to move the student toward the engine.

\subsection{Rules for the GRPO and REINFORCE comparison}
\label{app:prereg:rl}

These rules judge the GRPO and REINFORCE runs of Section~\ref{sec:rl-contrast}.

\paragraph{Recipe and learning test.}
The problems are those of DAPO-Math-17k~\citep{yu2025dapo} on which the untrained student solved 1 to 7 of
8 samples; GRPO's second update per step clips the probability ratio to $[0.8, 1.28]$. The plan first required each
method to pass a learning test with the standard engine (among its criteria, a held-out accuracy gain larger than the
run-to-run spread). In one run each at a learning rate of $10^{-6}$, both stayed healthy at steps 100 and 200 but did
not pass. The condition was dropped after those runs were read and before any run under mismatch.

\paragraph{Breakage, judged the same way for every method.}
Against its own method's standard run, a run is \emph{broken} by a held-out accuracy drop of at least 5
points with a problem-paired interval excluding zero and larger than the plan's run-to-run margin, an entropy or
length collapse confirmed on held-out samples, a non-finite loss or, for OPD, the primary rule. It is \emph{collapsed}
if, in addition, its reward is zero for five steps in a row, its responses stay at the length limit, or its entropy
falls below half its standard-engine level.

\paragraph{Score centering at one engine per method.}\label{app:rl-sc}
At one engine where each method broke, it was trained twice more: with score centering over the engine's 128
(REINFORCE) or 256 (GRPO) most likely tokens, and with a control that computes the centering term without applying it.
By rules fixed in advance, centering rescues a run if the centered run shows none of the uncorrected run's failures,
and the rescue is due to centering if the control still shows them. One run per side.

\emph{REINFORCE}, with Model Optimizer's FP8 recipe and the first and last layers in BF16: the run without
corrections lost held-out accuracy and its reward fell below a band built as in the collapse rule
(App.~\ref{app:prereg:rules}) from REINFORCE's standard run. The control shows both failures, the centered run neither;
its held-out accuracy is +60.98 points [+58.67, +63.32] above the uncorrected run's at step 100.

\emph{GRPO}, with NVFP4 (emulated) and the first layer's KV cache in BF16: the run without corrections
failed only by an entropy collapse (below its band from step 29, below half its standard-run level from step 72), which the control shows and the centered run does not. Against the run without corrections, the centered run is level
at step 100 ($-0.15$ points [$-1.62$, $+1.35$]), so the rules' stricter label, a rescue at no accuracy cost, is missed
by 0.025 points; it is +16.50 points [+14.70, +18.32] higher at step 200.

\subsection{Speed readings}
\label{app:prereg:speed}

These rules produce the speed numbers of Section~\ref{sec:speed}. They give no speed or memory number for an emulated
format and, as first written, no claim that quantization makes a step faster.

\begin{itemize}
  \setlength{\itemsep}{1pt}\setlength{\parskip}{0pt}
  \item \emph{Throughput.} Updates per hour from step 3, relative to the synchronous standard run on the
    same GPUs; an asynchronous run's final steps, which only finish rollouts already in flight, are left out.
  \item \emph{Large rollout batch.} Eight-step runs at 1{,}024 and 1{,}536 sequences per step, with no
    quality claim; a memory-saving gain in updates per hour is credited only at 1{,}536
    and only to vLLM's FP8 cache with the first layer's KV cache in BF16.
  \item \emph{Time to quality.} The combined run uses Model Optimizer's FP8 recipe with the first and last layers
    in BF16 and the largest staleness whose run passes the primary rule at step 100 (8 versions). It and the synchronous standard run save checkpoints every 25 steps; the
    targets are the latter's step-200 held-out KL and 90\% of its improvement from step 0 to step 200.
\end{itemize}

\subsection{Admission predictions}
\label{app:prereg:admit}

These predictions, fixed before training against the admission line of 0.41~nats/token, form the record of
Table~\ref{tab:admission} (Section~\ref{sec:shortcuts}). Eight held, and a forecast that vLLM's FP8 cache at its
default scale on every layer fails was hit. For every admitted engine, 1.5 times its estimate is at most 0.045; for
every rejected one, at least 0.61. So any line in between, including the lower end of our 7B student's budget,
0.184, gives the same record.

\subsection{One run per configuration}
\label{app:prereg:onerun}

Each configuration is trained once, except four trained twice, so every verdict is provisional; each
pair is reported as a range, never pooled. Repeats of the standard and combined runs, which only save checkpoints more often,
end at step 200 at a held-out KL of 0.047643 against 0.048103 (1.0\% lower) and 0.047602 against 0.048322 (1.5\%
lower). Repeats at 0.097 and 0.39~nats/token with a second rollout seed differ from their first runs by under 1 point
of relative held-out KL at steps 100 and 200. The synchronous standard run took 5.8 and 6.0 hours in two runs, the second
saving six more checkpoints.

\subsection{Experiments and their status}
\label{app:prereg:list}

Table~\ref{tab:app-experiments} lists every experiment; training runs last 200 steps unless stated.

\begin{table*}[!tp]
\centering
\caption{\textbf{Every experiment of the study.}}
\label{tab:app-experiments}
{\scriptsize\setlength{\tabcolsep}{3pt}%
\begin{tabular}{@{}p{0.25\textwidth}p{0.57\textwidth}p{0.14\textwidth}@{}}
\toprule
Experiment and question & What was run & Reported in \\
\midrule
\raggedright \emph{Budget.} How much mismatch does OPD tolerate? & \raggedright Standard run;
  weight noise at four levels, two run twice; four Model Optimizer engines; vLLM's FP8 cache, with and without
  the first layer's KV cache in BF16; staleness of 1, 2, 4 and 8 versions; a colocated asynchronous run & \raggedright
  Section~\ref{sec:budget} \tabularnewline
\raggedright \emph{Admission.} Can we tell before training which engines fit? & \raggedright
  Offline estimates; the 1.5B and
  Llama-3.2-3B students with the standard engine and three and two FP8 engines & \raggedright
  Sections~\ref{sec:shortcuts}, \ref{sec:students} \tabularnewline
\raggedright \emph{Speed.} What does the budget buy? & \raggedright Speed and memory
  benchmarks; eight-step runs at 1{,}024 and 1{,}536 sequences per step; four 30-step runs without the draft head;
  the combined run and time to quality & \raggedright Section~\ref{sec:speed} \tabularnewline
\raggedright \emph{GRPO and REINFORCE.} Do they break where OPD holds? & \raggedright Four of
  OPD's engines at two learning rates (the second set not read, by the authors' decision); score centering;
  under staleness and on the smaller students: running, not reported & \raggedright Section~\ref{sec:rl-contrast} \tabularnewline
\raggedright \emph{Mechanism.} Why does OPD tolerate mismatch? & \raggedright Gradient probe;
  error at later checkpoints; score centering; five offset runs; two truncated-importance-sampling runs; corrections at
  fixed prefixes (undetermined); prefix shift & \raggedright Section~\ref{sec:discussion},
  App.~\ref{app:probe} \tabularnewline
\raggedright \emph{Qwen3 pair.} Does the budget hold for a Qwen3 student? & \raggedright
  Qwen3-4B with a Qwen3-30B-A3B teacher: training loss at step 50 still 0.854 of its start, above the 0.80
  limit; stopped & \raggedright This table; not retried \tabularnewline
\bottomrule
\end{tabular}}
\end{table*}

\subsection{Decisions made with data in hand}
\label{app:prereg:post}

These decisions, taken with data in hand, change what the paper reports.

\begin{itemize}
  \setlength{\itemsep}{1pt}\setlength{\parskip}{0pt}
  \item \emph{The pilot.} The study's hypotheses, its 200-step length, its production recipes and four constants
    (the offset's size, the admission line and its safety margin, the largest noise level) were chosen with a 100-step pilot's data (App.~\ref{app:pilot}).
  \item \emph{GRPO and REINFORCE.} The learning condition was dropped after the standard runs were read; the
    engines were chosen after OPD's results at them, and the noise level after REINFORCE collapsed at it under an
    earlier recipe.
  \item \emph{Later runs.} After the fifteen original runs of Table~\ref{tab:budget} were read, we added
    second runs at 0.097 and 0.39, a level at 0.184~nats/token, and the correction, prefix-shift and
    first-layer measurements, each with readings fixed before its jobs, except that the first-layer predictions knew an
    earlier, unrecorded measurement and one correction check was loosened from a relative error of $10^{-4}$ to
    $10^{-3}$ after data.
  \item \emph{Readings fixed after some data.} The admission record's window (steps 1--100) was fixed after
    the first Model Optimizer run's data; steps 1--200 give the same verdicts. The drift test's leveling-off reading was
    fixed knowing the values at $\sigma=0.15$, before the other two runs were read. The 1{,}536-sequence set was added
    after the benchmark without training, before the first large-batch set started.
  \item \emph{The combined run's engine} was changed before the run from the every-layer FP8 recipe, which
    the offline estimates put outside the budget, to the same recipe with the first and last layers in BF16.
  \item \emph{The check without the draft head} was added after every other speed result was read. By
    the authors' decision after its data, its Model Optimizer FP8 result is reported as a gain in updates per hour at the training batch,
    which the plan had excluded.
  \item \emph{The run with weights up to 4 versions old and FP8} was added after the 8-version run's time to
    quality was known; it ends 0.8\% above the synchronous run's final held-out KL, so Section~\ref{sec:speed} reports
    its time to the 90\% target.
\end{itemize}

%% file: sections/a-engines.tex
\section{Rollout Engines}
\label{app:recipes}
\label{app:quant}

This appendix defines the rollout engines of Section~\ref{sec:setup-samplers} and the offline estimate of Section~\ref{sec:setup-admission}.

\subsection{Quantized rollout engines}

Table~\ref{tab:app-recipes} lists the quantized engines of Sections~\ref{sec:budget} and~\ref{sec:shortcuts}: Model Optimizer's recipes~\citep{nvidia2026modelopt} and the FP8 attention cache of vLLM~\citep{kwon2023pagedattention}. FP8 is E4M3 throughout.

\begin{table}[!tb]
\centering
\caption{Quantized rollout engines. All run on vLLM~0.19.1's own kernels unless marked emulated. FP8 caches use calibrated scales unless marked scale 1. $^\dagger$After a one-line fix to vLLM~0.19.1's W4A8 scale conversion.}
\label{tab:app-recipes}
{\footnotesize\setlength{\tabcolsep}{3pt}%
\begin{tabular}{@{}p{0.50\columnwidth}p{0.46\columnwidth}@{}}
\toprule
Engine & Weights / activations / cache \\
\midrule
\multicolumn{2}{@{}l}{\emph{Trained}} \\
\raggedright vLLM FP8 cache & BF16 / BF16 / FP8, scale 1 \\
\raggedright \quad first layer's KV cache in BF16 & BF16 / BF16 / FP8, scale 1 \\
\raggedright Model Optimizer FP8 & FP8 / FP8 / FP8 \\
\raggedright \quad first and last layers in BF16 & FP8 / FP8 / FP8 \\
\raggedright NVFP4, emulated & FP4 / FP4 / FP8 \\
\raggedright \quad first layer's KV cache in BF16 & FP4 / FP4 / FP8 \\
\addlinespace
\multicolumn{2}{@{}l}{\emph{Offline estimate only} (App.~\ref{app:quant:offline}; Figure~\ref{fig:shortcuts})} \\
\raggedright INT8 SmoothQuant; INT4 AWQ; emulated & as shipped / as shipped / BF16 or FP8 \\
\raggedright W8A16: Model Optimizer's FP8 weights, stored in BF16 & FP8 / BF16 / BF16 or FP8 \\
\raggedright W4A8: llm-compressor AWQ, INT4 in groups of 128$^\dagger$ & INT4 / FP8 / BF16 or FP8 \\
\bottomrule
\end{tabular}}
\end{table}

\paragraph{Emulated formats.}
NVFP4~\citep{alvarez2025nvfp4} needs Blackwell GPUs, so it is emulated: tensors are rounded to its grid with Model Optimizer's scales and used in BF16, checked bit for bit against Model Optimizer's fake quantization on 21 captured layers. Emulated formats give mismatch and quality only.

\paragraph{Calibration set.}
Calibrated scales come from 512 untrained-student responses (temperature 1) to 256 DAPO-Math-17k training problems (seed 20261008), disjoint from all held-out and evaluation sets. Activation and cache scales stay fixed; new weights are re-quantized at every sync.

\paragraph{The default FP8 cache.}
Without cache scales in the checkpoint, vLLM stores the cache in FP8 at scale 1, coarse for the 7B student's large first-layer keys (per-head RMS 10 to 43). A vLLM option keeps listed layers' cache in BF16.

\subsection{Offline estimate of the mismatch}
\label{app:quant:offline}

The offline estimate runs on one GPU before training. The engine under test samples 2 responses to each of 16 prompts (problems 0--7 of AIME24 and of MATH-500; temperature 1, top-$p$ 1, up to 16{,}384 tokens, seed 777). The estimate is the mismatch of Eq.~\eqref{eq:dose} on these tokens, with the standard engine in place of the student's $p_\theta$ and a 95\% bootstrap interval over prompts. The standard engine against itself gives 0.00057 [0.00044, 0.00070] nats/token.

\paragraph{First-layer share.}
To test whether the first layer dominates on another family (Section~\ref{sec:shortcuts}), the share $(A-B)/(A-F)$ compares the full recipe ($A$), the recipe with the first layer's KV cache in BF16 ($B$) and the standard engine ($F$). On Qwen3-4B~\citep{yang2025qwen3}, the score-centering code's FP8 recipe~\citep{marek2026code} (emulated) gives 0.0048 [0.0032, 0.0074] nats/token and a share of $-0.18$ [$-0.97$, 0.36], against 0.515 and 0.968 for the same recipe on the 7B student.

\subsection{Why the first layer's keys carry the mismatch}
\label{app:firstlayer}

This measurement on the untrained 7B student finds which first-layer tensor carries the FP8 cache's mismatch (Section~\ref{sec:shortcuts}). Table~\ref{tab:app-firstlayer} gives the offline estimate, the only unit compared with the budget, and the exact KL at the 1{,}024 fixed prefixes of App.~\ref{app:frozen}.

\begin{table}[!tbp]
\centering
\caption{The first layer's FP8 cache, one tensor at a time (untrained 7B student; one measurement per row; emulated at scale 1 unless marked real). Offline estimate in nats/token, with 95\% intervals over prompts. Fixed prefixes: KL relative to keys and values. $^a$Ratio of emulated to real. $^b$Share of the every-layer cache's KL that a BF16 first layer removes.}
\label{tab:app-firstlayer}
{\scriptsize\setlength{\tabcolsep}{3pt}%
\begin{tabular}{@{}lrr@{}}
\toprule
In FP8 & Offline estimate & Fixed prefixes \\
\midrule
Nothing (standard engine) & 0.0006 & -- \\
Values & 0.0005 & 0.02\% \\
Keys & 0.51 [0.45, 0.57] & 99.95\% \\
Keys and values & 0.62 & 100\% \\
vLLM's first-layer cache, real & 0.55 & 1.03 [0.96, 1.10]$^a$ \\
Every layer but the first, real & 0.0037 & 99.9\%$^b$ \\
\bottomrule
\end{tabular}}
\end{table}

\paragraph{The keys carry it.}
Alone, the first layer's keys give 82.5\% [73.7, 91.4] of the keys-and-values offline estimate.

\paragraph{Why keys, not queries.}
The first layer's keys share a large common component, so FP8 adds a different error to each score in a query's row of attention scores. A query's error, multiplied by these nearly alike keys, shifts that whole row almost equally, which the softmax ignores. Rounding the queries, as vLLM does, has 4.9\% of the keys' effect on the attention weights.

\paragraph{The real kernel.}
A tensor-level check confirmed bit for bit that vLLM's FP8 cache rounds these tensors as the emulation does.

\paragraph{Missed predictions.}
Extra mantissa bits for the keys lower the KL more steeply than predicted, and no measure we computed ranks the four first-layer key--value heads as their KL does.

\subsection{Rollout engine with weight noise}
\label{app:noise}

\paragraph{Construction.}
At each weight sync the engine adds $\sigma s_w\varepsilon_w$ to each of the 196 projection matrices $W$ in the transformer blocks; $s_w$ is the standard deviation of $W$ at the first sync, and $\varepsilon_w\sim\mathcal N(0,I)$ is seeded by a hash of seed 20261001 and the matrix's name $w$. The offset $n=(\sigma s_w\varepsilon_w)_w$ is thus fixed across syncs and replicas, and unseen by the trainer.

\paragraph{Direction.}
The unit direction $\hat n=n/\|n\|$ does not depend on $\sigma$; the 200 random directions of App.~\ref{app:prereg:rules} use seeds 1 to 200 instead, with the same $s_w$.

\paragraph{Levels.}
The levels $\sigma=0.0665$, 0.15 and 0.325 ran at 0.055, 0.097 and 0.39~nats/token (Table~\ref{tab:budget}). The level $\sigma=0.234$ was added after the other runs had been read, to fall between the passing and failing levels; it ran at 0.184 and ended 0.5\% better at step 200, against a predicted 3\% to 12\% worse.

\subsection{Staleness}
\label{app:staleness}

This subsection defines the stale-weight engines (Sections~\ref{sec:budget} and~\ref{sec:speed}) and one never refreshed (Section~\ref{sec:discussion}).

\paragraph{Fully asynchronous runs.}
Rollout and training use separate GPUs; responses may be up to $k\in\{1,2,4,8\}$ versions older than the trainer. All $k$ share one split of the synchronous reference's total GPUs: the trainer on two of the student node's four, two rollout replicas on the other two, and the teacher's four replicas on a second node. With no run at $k=0$, the $k=1$ run shows the asynchronous machinery's cost.

\paragraph{Colocated run.}
Trainer and rollout share the GPUs. After one warm-up batch, a token is trained only if it is at most two weight versions old. A step starts once 8 of the 16 prompt groups in flight finish; interrupted generations resume under the new weights.

\paragraph{An engine whose weights are never refreshed.}
\label{app:frozen}
Semi-OPD~\citep{zhao2026semiopd} and Lightning OPD~\citep{wu2026lightning} rescore rollouts drawn once from the untrained student with the current student, without importance weights. Their mismatch at step $n$ is the student's change since step 0, $\mathrm{KL}(p_{\theta_0}\,\|\,p_{\theta_n})$, computed exactly on the 1{,}024 fixed prefixes of Section~\ref{sec:mechanism}: 16 uniform positions in one untrained-student response to each of 64 held-out problems (temperature 1, capped at 4{,}096 tokens). It is 0.082~nats/token at steps 100 and 200; the checkpoints come from standard runs, and a student trained on fixed rollouts may change differently.

%% file: sections/a-probe.tex
\section{The Gradient Probe}
\label{app:probe}

This appendix holds the analyses behind the discussion of why OPD tolerates mismatch (Section~\ref{sec:mechanism}).

\paragraph{Setup.}
At the untrained student, the probe computes exactly the expected loss gradient of OPD (the negative of the update of Section~\ref{sec:background}, so $G_{\mathrm{bias}}=-g^z_{\mathrm{bias}}$ of Eq.~\eqref{eq:mech-error}) with the advantage of Eq.~\eqref{eq:bg-adv} under the weight-noise engines at $\sigma=0.0665$ and 0.15, and its drift part ($G_{\mathrm{drift}}$ below). At a prefix, $p$, $p_T$ and $q$ are the student's, teacher's and engine's next-token distributions, $z$ the student's logits and $A_v=\log p_T(v)-\log p_v$. Token $y$'s loss $\ell(y)=-A_y\log p_y$, with $A_y$ fixed, has $\nabla_z\ell(y)=-A_y(e_y-p)$ ($e_y$ one-hot; $\circ$, $\cdot$: element-wise and inner products):
\begin{equation}
\begin{aligned}
  G_{\mathrm{KL}}    &= \mathrm{E}_{y\sim p}\,\nabla_z\ell(y) = -\bigl(p\circ A-(p\cdot A)\,p\bigr),\\
  G_{q}              &= \mathrm{E}_{y\sim q}\,\nabla_z\ell(y) = -\bigl(q\circ A-(q\cdot A)\,p\bigr),\\
  G_{\mathrm{bias}}  &= G_q-G_{\mathrm{KL}},\\
  G_{\mathrm{drift}} &= -(q\cdot A)\,(q-p).
\end{aligned}
\label{eq:app-fields}
\end{equation}
$G_{\mathrm{KL}}=\nabla_z\mathrm{KL}(p\|p_T)$ is the on-policy target, $G_{\mathrm{bias}}$ what the engine adds, and $G_{\mathrm{drift}}$ the part score centering removes~\citep{marek2026score}. Backpropagating $\langle\mathrm{stopgrad}(G_t),z_t\rangle$, averaged over response tokens $t$, gives each $G$'s parameter gradient $g$; $g_q=g_{\mathrm{KL}}+g_{\mathrm{bias}}$, $r_{\mathrm{bias}}=\|g_{\mathrm{bias}}\|/\|g_{\mathrm{KL}}\|$, and $\hat n$ is the unit direction of the noisy engine's weight offset (App.~\ref{app:noise}).

The prefixes are all positions of \ProbeTraces{} responses of the untrained student (\ProbeRows{} tokens); $q$ enters only through the next-token distributions. The closed form matches autograd (relative error \ProbeIdentityErr{} in fp32).

\begin{table}[!tbp]
\centering
\caption{Gradient probe, weight-noise engines. Probe mismatch: $\mathrm{KL}(q\|p)$ at the untrained student on its own responses; in training these engines average 0.055 and 0.097 (Table~\ref{tab:budget}).}
\label{tab:app-probe}
{\scriptsize\setlength{\tabcolsep}{3pt}%
\begin{tabular}{@{}lrr@{}}
\toprule
 & $\sigma=0.0665$ & $\sigma=0.15$ \\ %
\midrule
Probe mismatch (nats/token) & \ProbeMMNmid & \ProbeMMNhigh \\
Token mean of $\mathrm{E}_q[A]$ & \ProbeEqANmid & \ProbeEqANhigh \\
\addlinespace
$r_{\mathrm{bias}}$ & \ProbeRbiasNmid & \ProbeRbiasNhigh \\
Angle between $g_q$ and $g_{\mathrm{KL}}$ (degrees) & \ProbeAngleNmid & \ProbeAngleNhigh \\
\addlinespace
$\langle g_{\mathrm{bias}},\hat n\rangle$ & \ProbeBiasProjNmid & \ProbeBiasProjNhigh \\
\bottomrule
\end{tabular}}
\end{table}

\paragraph{Reading.}
The mismatched update points in nearly the right direction (Table~\ref{tab:app-probe}): at $\sigma=0.0665$ and 0.15 the bias is \ProbeRbiasPctNmid\% and \ProbeRbiasPctNhigh\% of $\|g_{\mathrm{KL}}\|$, and $g_q$ lies within \ProbeAngleNmid{} and \ProbeAngleNhigh{} degrees of $g_{\mathrm{KL}}$. The bias projects positively on $\hat n$, so a gradient step moves the student away from the engine along $\hat n$, as a negative drift coefficient $\mathrm{E}_q[A]$ predicts.

\paragraph{The error at later checkpoints.}
The error shrinks in training; the update shrinks faster. On the 1{,}024 fixed prefixes of App.~\ref{app:frozen}, each checkpoint's engine adds weight noise scaled to a mismatch of 0.05, 0.2 or 0.4~nats/token (within 10\%). The reading fixed in advance is the ratio of the error of Eq.~\eqref{eq:mech-error} (summed per-prefix norm) at step 200 of the standard run to that at the untrained student: 0.130 [0.108, 0.153], 0.130 [0.110, 0.151] and 0.109 [0.094, 0.126] (95\% bootstrap intervals over problems). At a mismatch of 0.2, the error-to-update ratio ($r_{\mathrm{bias}}$ on 64 prefixes, a descriptive reading) is 0.93 at the untrained student and 1.88 at step 200.

\subsection{Corrections at a fixed prefix}
\label{app:corrections}

This subsection tests whether per-token corrections, which did not help OPD in training (Section~\ref{sec:mechanism}), reduce the error at a fixed prefix, and what reweighting whole responses costs.

\paragraph{Measurement.}
For tokens drawn from the engine, we compare five per-token estimators with the exact on-policy update: plain OPD, score centering (exact, and on the engine's top 128 tokens as in our runs), and importance weights $p_\theta/q$, plain and truncated at 2. The error is the squared distance for a batch of about 480{,}000 tokens, on the 1{,}024 fixed prefixes (64 in parameter space), at the untrained student and steps 100 and 200 of standard runs, for $\sigma=0.0665$, 0.15 and 0.325.

\paragraph{Readings.}
At $\sigma=0.15$ and steps 100 and 200, the 95\% intervals of the error ratio to plain OPD, at 64 prefixes, leave it undetermined whether truncated weights or score centering on 128 tokens reduce the error. Two of seven predictions, fixed in advance, about the corrections' per-token variance held.

\paragraph{Whole-response reweighting.}
Only reweighting whole responses by $p_\theta/q$ reaches the prefix shift. Its expected effective fraction is at most $\exp(-\mathrm{KL}(p_\theta\|q))$ over whole responses; approximating that KL by 7{,}538 tokens (the mean response length) times the mean per-token KL at the fixed prefixes gives $10^{-27}$ or smaller even at $\sigma=0.0665$: about $10^{27}$ responses per effective sample, a population estimate, not a bound.

\subsection{A teacher-anchored objective at a fixed prefix}
\label{app:anchor}

This subsection proves the stationarity statement of Section~\ref{sec:mechanism} at one prefix, where $q$, $p_\theta$ and $p_T$ are the engine's, trainer's and teacher's next-token distributions. With $A_v=\log p_T(v)-\log p_\theta(v)$ held fixed, as in verl, and a probability ratio of 1, the expected loss gradient is $G_q=-\mathbb{E}_q[A\,\nabla_\theta\log p_\theta]$, in parameter space.

\paragraph{Proposition.} For fixed $q$, let $J_q(\theta)=\tfrac12\,\mathbb{E}_{v\sim q}[A_v^2]$, now with $A$ a function of $\theta$. (i) $\nabla_\theta J_q=G_q$. (ii) At $p_\theta=p_T$, every sampled token's update is zero, for every engine. (iii) There, the Hessian of $J_q$ in the prefix's logits $z$ is $B^\top\mathrm{diag}(q)\,B$, $B=I-\mathbf{1}p_T^\top$: positive semidefinite and, if all $q_v>0$, positive definite except in the direction that adds a constant to every logit.

\paragraph{Proof.} (i) $q$ and $p_T$ do not depend on $\theta$, so $\nabla_\theta J_q=\mathbb{E}_q[A\,\nabla_\theta A]=G_q$. (ii) Every $A_v=0$. (iii) As $\nabla_z\log p_\theta(v)=e_v-p_\theta$, differentiating $\nabla_zJ_q=-\sum_vq_vA_v(e_v-p_\theta)$ again at $A=0$ leaves $\sum_vq_v(e_v-p_T)(e_v-p_T)^\top=B^\top\mathrm{diag}(q)\,B$, and $u^\top B^\top\mathrm{diag}(q)\,Bu=\sum_vq_v(u_v-p_T^\top u)^2\ge0$, which, if all $q_v>0$, is zero only for constant $u$. $\square$

\paragraph{Scope.} The squared log ratio is the ``k2'' estimator of the KL~\citep{schulman2020klapprox}. The result needs the trainer's advantage: with the engine's, $\log p_T(v)-\log q(v)$, the update at $p_\theta=p_T$ is zero only if $q=p_T$. The result holds at one prefix with the engine fixed, so it bounds neither the cost of mismatch nor the budget.

\subsection{The prefix shift}
\label{app:shift}

This subsection splits the update's error into the per-token error $T$, the only part per-token corrections act on, and the prefix shift $S$ (Figure~\ref{fig:prefix-shift}).

\paragraph{Measurement.}
For 64 held-out problems, four responses each come from the standard engine, an independent redraw (a placebo, true shift zero), and the engines with weight noise at $\sigma=0.15$ (passes in training) and 0.325 (fails). At the untrained student and step 200 of the standard run, the expected per-token update is exact at every prefix. Against the on-policy update at the standard engine's prefixes, the error splits exactly into $T$, the error at the engine's own prefixes, and $S$, the on-policy update's change when the prefixes move to the engine's. Sizes are cross-fitted over response halves. Two FP8-cache engines were also measured (the every-layer one's $T$ was flagged); only the weight-noise sizes are reported.

\paragraph{Readings.}
For weight noise, the shift meets the rule fixed in advance: it exceeds the placebo, carries a fifth or more of the error's squared size, and from the passing to the failing level grows (paired interval excluding zero) by a larger factor than $T$. At the untrained student, relative to the on-policy update, $S$ rises from 0.098 at $\sigma=0.15$ to 0.517 at 0.325 (5.3 times) and $T$ from 0.198 to 0.417 (2.1 times), both increases with intervals excluding zero. The split depends on which change is made first: at the untrained student, $S$'s factor is also the larger in the reverse order (21.6 against $T$'s 12.8) and in the symmetric split (15.2 against 8.0); at step 200 only the order above favors $S$ (7.8 against 5.9).

%% file: sections/a-speedrepro.tex
\section{Real-Kernel Speed Benchmark}
\label{app:speedbench}

This appendix gives the real-kernel evidence for Section~\ref{sec:speed}.

\paragraph{Question and setup.}
Does a rollout with vLLM's own low-precision kernels shorten a step at our training shape, 16 sequences
per replica? A vLLM~0.19.1 benchmark without training timed BF16, FP8 weights in verl's recipe, vLLM's
FP8 cache and both, with and without the draft head (App.~\ref{app:repro}), at 16, 64, 128 and, descriptively, 256
per replica.

\paragraph{Rule, fixed in advance.}
A configuration within a fixed mismatch limit passes at a batch size if its speedup in batch time, in the
median and on every paired prompt set, reaches the threshold against BF16 with and without the draft head: 1.15 at 16 per replica and
1.30 at 64 and 128, each raised by BF16's job-to-job spread.

\paragraph{Result.}
None passed: within the limit, the best speedups were \GzeroBest{} at 16, 64 and 128 per replica, against
raised thresholds of \GzeroThr{}. The cache never filled there (peak use at most \GzeroKVuseMax, \GzeroPreemptMax{}
preemptions); BF16's cache first filled at 256. Model Optimizer's recipe was not in this benchmark; Table~\ref{tab:speed-64}
gives its gain in training. The FP8 cache's benefit is memory: with the first layer's KV cache in BF16 it holds
\GzeroKVcapKV$\times$ as many tokens as BF16's, at a real-kernel mismatch of
\GzeroMismatch{} nats/token.

\paragraph{What a faster rollout could buy.}
With the draft head, generation and teacher scoring take $f=\TimeGenCleanPct\%$ of a mean \StepSCleanMean\,s
step in the pilot's three 100-step standard runs (App.~\ref{app:pilot}); without it, 76\% in this study
(Table~\ref{tab:speed-64}). A rollout $s$ times faster raises the step rate by at most
$1/((1-f)+f/s)$: \AmdahlGzerob$\times$ for $s=1.25$, never above \AmdahlCeiling$\times$.

\paragraph{Memory test, fixed in advance.}
At 1{,}536 sequences per step the BF16 cache fills. Memory is shown if the measured ratio $R$ of updates
per hour beats what FP8's speed alone predicts, $R_{\mathrm{np}}=1/((1-f)+f/s_{\mathrm{np}})$, by
$G=R/R_{\mathrm{np}}\ge1.05$; $f$ is this BF16 run's generation and teacher-scoring share (0.42), $s_{\mathrm{np}}$ the FP8
cache's generation speedup without training at this batch, with memory to spare (1.13). For the FP8 cache,
$R_{\mathrm{np}}=1.051$ and $R=1.111$, so $G=1.057$: memory is shown ($G=1.007$ at 1{,}024 per step). Model Optimizer
FP8 gets no verdict: no $s_{\mathrm{np}}$ was measured for it. One run each; repeated runs of one configuration have
differed by up to 3.6\% in run time.

\paragraph{Without the draft head.}
Table~\ref{tab:speed-64} reruns four configurations for 30 steps without the draft head: vLLM's FP8 cache,
Model Optimizer FP8 and weights up to 4 versions old give 1.002, 1.266 and 2.008 times BF16's updates per hour. With
the head they give 0.989, 1.079 and 1.643 over the same steps 3--30, as over steps 3--200 (0.990, 1.082, 1.624), so the
head, not the shorter window, explains the difference.

\section{The Pilot}
\label{app:pilot}

A 100-step pilot on our 7B student with emulated quantization preceded this study; its results are not
reported. Its eight quantized engines trained at mismatches up to 1.47 times their offline estimates,
which the admission check's margin of 1.5 covers (Section~\ref{sec:setup-admission}). Six outputs are fixed
inputs here: the step-time breakdown of its standard runs (App.~\ref{app:speedbench}); their mean held-out
KL, the step-100 reference (App.~\ref{app:prereg:rules}); the admission line of 0.41~nats/token
(Section~\ref{sec:admission-method}); those ratios; two checkpoints of one standard run
(Section~\ref{sec:mechanism}); and its weight-noise run at $\sigma=0.15$, the OPD run that GRPO and REINFORCE at that
engine are compared with at launch and at step 100 (Section~\ref{sec:rl-contrast}).

\section{Reproducibility}
\label{app:repro}
\begin{table*}[!t]
\caption{\textbf{Every OPD run on the mismatch axis} (one 200-step run per row). $D$: mismatch in
training, mean over steps 1--200 (nats/token), which names noise levels in the main text (0.097: $\sigma=0.15$). Accuracy: composite change at step 200. Intervals are over problems within a run.}
\label{tab:budget}
\centering
\input{tables/budget_runs}
\end{table*}

\paragraph{Code and data.}
The code is at \url{https://github.com/benyucong/train-inference-mismatch-opd/}.
It holds the engines and their emulation, the training launcher, the offline estimate, the speed benchmark,
the gradient probe, the prefix-shift measurement, the held-out evaluation and the preparation of every dataset. The
plan, the per-step logs, the analyzers' outputs and the scripts that write the numbers and figures will be released
after the review.

\paragraph{Software and hardware.}
The trainer is verl~\citep{sheng2025hybridflow} at commit \texttt{fbb4b3a8} with FSDP~\citep{zhao2023fsdp};
vLLM~0.19.1~\citep{kwon2023pagedattention} with PyTorch~2.10 runs rollouts and the teacher. A synchronous OPD run
uses two nodes of four GH200 GPUs, one for the trainer with four rollout replicas and one for the teacher; GRPO and
REINFORCE use one. The driver is 595.91.07 with CUDA 13.2; the second runs at 0.097 and 0.39 (Section~\ref{sec:budget}), and
most likely the FP8 run with weights up to 4 versions old (Figure~\ref{fig:speed-async}), used 615.71.09.

\paragraph{Models and data.}
The 7B student and teacher use revisions \texttt{916b56a4} and \texttt{fcef970f}. Every OPD and RL run
on this pair decodes rollouts with a public EAGLE-3 draft head~\citep{li2025eagle3} and 3 draft tokens,
except the memory tests and the 30-step runs of Table~\ref{tab:speed-64}; offline estimates, evaluations and the smaller
students use none. Speculative decoding leaves the rollout engine's distribution
unchanged~\citep{leviathan2023speculative}, and every speed comparison pairs runs with the same setting. The head's published configuration omits
\texttt{max\_position\_embeddings}, so vLLM builds a rotary cache for only 2{,}048 positions; we set the field to
131{,}072. Training prompts are
DAPO-Math-17k~\citep{yu2025dapo}; the held-out set is AIME24~\citep{maa2024aime} and
MATH-500~\citep{hendrycks2021math,lightman2024verify}, including 38 problems flagged as overlapping the
training pool.

\paragraph{Training configuration.}
Each OPD run takes 200 steps of 8 prompts $\times$ 8 samples at temperature 1 without top-$p$ or top-$k$,
at most 16{,}384 response tokens and one AdamW step per batch at a constant learning rate of $10^{-6}$ (weight decay
0.01, gradient clipping at 1.0). The loss is verl's k1 estimator in policy-gradient form, with the advantage of
Eq.~\eqref{eq:bg-adv}; task rewards, KL penalties, entropy bonus and rollout importance corrections are off.

\paragraph{Seeds, evaluation and compute.}
Seeds are 1 for the data order, 20261001 for the weight noise and, for rollouts, 42 (43 in the second
runs at 0.097 and 0.39) plus $r$ for replica $r$; the engine is not deterministic. Every checkpoint is evaluated with
the standard engine at temperature 1, up to 16{,}384 tokens, with evaluation seeds fixed per problem and sample and
shared by all runs; a rule checks the last boxed answer. A 200-step synchronous run took 5.5 to 6.0 hours on
eight GPUs. Scripts write most numbers in the gradient probe and the benchmark result
(App.~\ref{app:probe} and~\ref{app:speedbench}); the per-run tables were copied by hand.

%% file: tables/budget_runs.tex
{\footnotesize\setlength{\tabcolsep}{3.5pt}%
\begin{tabular}{@{}lrrrrl@{}}
\toprule
 & Mismatch & \multicolumn{2}{c}{Held-out KL vs.\ the standard run (\%) [95\% interval]} & Accuracy & Primary rule \\
\cmidrule(lr){3-4}
Rollout engine & level $D$ & Step 100 & Step 200 & (points) & (100 / 200) \\
\midrule
Standard engine (reference) & 0.0008 & (KL 0.0573) & (KL 0.0481) & (60.93\%) & -- \\
\addlinespace
\multicolumn{6}{@{}l}{\emph{Asynchronous rollouts, standard engine}} \\
Colocated, at most 2 versions old & 0.0009 & $-0.95$ [$-2.85$, $+0.95$] & $-0.16$ [$-2.61$, $+2.31$] & $-0.93$ & pass / pass \\
Staleness of 1 model version & 0.0009 & $-1.93$ [$-4.27$, $+0.39$] & $-2.72$ [$-5.21$, $-0.30$] & -- & pass / pass \\
Staleness of 2 model versions & 0.0011 & $-1.35$ [$-3.59$, $+0.89$] & $-0.82$ [$-3.50$, $+1.89$] & -- & pass / pass \\
Staleness of 4 model versions & 0.0016 & $-1.21$ [$-3.37$, $+0.92$] & $+1.41$ [$-1.21$, $+4.00$] & -- & pass / pass \\
Staleness of 8 model versions & 0.0028 & $-0.95$ [$-3.18$, $+1.30$] & $+0.29$ [$-2.36$, $+2.93$] & -- & pass / pass \\
\addlinespace
\multicolumn{6}{@{}l}{\emph{Quantized engines, first layer's KV cache kept in BF16}} \\
vLLM's own FP8 cache & 0.0063 & $-0.01$ [$-2.14$, $+2.07$] & $-1.53$ [$-4.06$, $+1.02$] & $+0.71$ & pass / pass \\
Model Optimizer FP8$^{\ddagger}$ & 0.0099 & $-1.60$ [$-3.42$, $+0.24$] & $-0.45$ [$-3.09$, $+2.21$] & $+0.88$ & pass / pass \\
\quad plus a staleness of 8 versions & 0.0118 & $-1.79$ [$-4.19$, $+0.63$] & $+0.45$ [$-2.40$, $+3.32$] & -- & pass / pass \\
Model Optimizer NVFP4$^{\ast}$ & 0.0485 & $-0.01$ [$-2.06$, $+2.06$] & $-5.04$ [$-7.51$, $-2.57$] & $+1.18$ & pass / pass \\
\addlinespace
\multicolumn{6}{@{}l}{\emph{Weight noise}} \\
$\sigma=0.0665$ & 0.0547 & $+5.27$ [$+3.08$, $+7.50$] & $+6.35$ [$+3.52$, $+9.15$] & $-0.30$ & pass / pass \\
$\sigma=0.15$ & 0.0968 & $+0.48$ [$-1.47$, $+2.52$] & $+0.26$ [$-2.31$, $+2.85$] & $+0.58$ & pass / pass \\
\quad second run$^{\P}$ & 0.0968 & $+0.07$ [$-1.87$, $+2.06$] & $-0.53$ [$-3.29$, $+2.17$] & $-1.37$ & pass / pass \\
$\sigma=0.234$ & 0.1841 & $+2.89$ [$+0.68$, $+5.16$] & $-0.51$ [$-2.98$, $+1.97$] & $+0.64$ & pass / pass \\
$\sigma=0.325$ & 0.3931 & $+19.54$ [$+16.97$, $+22.11$] & $+23.28$ [$+19.92$, $+26.60$] & $-0.66$ & pass / \textbf{fail} \\
\quad second run$^{\P}$ & 0.3928 & $+19.51$ [$+17.09$, $+21.96$] & $+23.69$ [$+20.50$, $+26.89$] & $-1.41$ & pass / \textbf{fail} \\
\addlinespace
\multicolumn{6}{@{}l}{\emph{Quantized engines, every layer}} \\
Model Optimizer NVFP4$^{\ast}$ & 0.5633 & $+41.87$ [$+39.22$, $+44.62$] & $+59.66$ [$+55.92$, $+63.47$] & $-5.21^{\S\dagger}$ & \textbf{fail} / \textbf{fail} \\
Model Optimizer FP8 & 0.6372 & $+40.80$ [$+38.02$, $+43.59$] & $+57.96$ [$+54.31$, $+61.75$] & $-3.41^{\S}$ & \textbf{fail} / \textbf{fail} \\
vLLM's own FP8 cache & 0.6803 & $+39.55$ [$+36.89$, $+42.19$] & $+53.92$ [$+50.08$, $+57.83$] & $-3.63^{\S}$ & \textbf{fail} / \textbf{fail} \\
\bottomrule
\end{tabular}\par\vspace{3pt}%
\parbox{0.98\linewidth}{\raggedright $^{\ast}$ Emulated arithmetic: a quality result only, with no memory or speed
number. $^{\ddagger}$ First and last layers in BF16. $^{\S}$ The 95\% interval of the accuracy change
excludes 0 (in no other row). $^{\dagger}$ The only accuracy failure by our definition (a drop of at least 5 points).
$^{\P}$ A second run of the configuration above with another rollout sampling seed (43 instead of 42); the
two runs are a range, not pooled. ``--'': no accuracy evaluation. References at each step: App.~\ref{app:prereg:rules}. For the staleness rows and vLLM's own cache,
the interval is that of the paired difference in nats per token over the reference's mean.}}

%% file: sections/a-tables.tex
\section{Full Result Tables}
\label{app:tables}

These tables back Sections~\ref{sec:budget} to~\ref{sec:discussion}: the
budget (Table~\ref{tab:budget}), the admission check (Table~\ref{tab:admission}), speed (Tables~\ref{tab:speed-async}
to~\ref{tab:speed-ttq}, Figure~\ref{fig:speed-steptime}) and changes to OPD's update (Table~\ref{tab:mech-tests}).
Every run uses our 7B student unless a row names another. Verdicts are provisional: one run per configuration,
except four configurations trained twice (App.~\ref{app:prereg:onerun}).

\begin{table*}[p]
\vspace{-6pt}%
  \caption{\textbf{The admission check's record} (one 200-step OPD run per row). \emph{Estimate}: offline,
  before training. \emph{Training}: mean mismatch over steps 1--100. \emph{Ratio}: training over
  estimate, predicted between $1/1.5$ and 1.5 ($^\ast$outside). \emph{Held-out KL}: against the standard run at the
  same step (first six rows at step 100: the mean of three 100-step standard runs; otherwise the pair's 200-step
  standard run). \emph{Accuracy}: change at step 200 ($^\dagger$fails the accuracy rule, App.~\ref{app:prereg:rules}).
  \emph{Predicted}: the check's verdict, fixed before the run (none: no prediction). Emulated: quality only. $^a$A forecast, not a fixed prediction. $^b$The estimate omits staleness; the prediction used the
  recipe's admission and the mismatch of staleness alone. $^c$No offline estimate exists; the mean mismatch of a 5-step test run.
  $^d$Not shown to fail: the smallest detectable rise is 14 to 20\%, and the
  interval reaches $+27.3\%$. $^e$Recorded before the run without an estimate; not one of the eight.}
  \label{tab:admission}
  \centering
  \input{tables/admission}
\vspace{-3pt}%
\centering
\begin{minipage}{0.86\textwidth}\centering
\caption{\textbf{Fully asynchronous OPD by staleness} (200 steps, eight GPUs; one run per row).
Staleness $k$: rollout weights up to $k$ versions older than the trainer's (exceeded by one version on one
step at $k=2$, three at $k=8$ and one at $k=8$ with FP8). Held-out KL: against the synchronous standard run at the same step. \emph{Trainer
waits}: seconds per step waiting for generation and teacher scoring. FP8: Model Optimizer's recipe, first and last layers
in BF16; the run at $k=4$ with it saves a checkpoint every 25 steps and most likely ran on a newer GPU driver
(App.~\ref{app:repro}). Runs use the EAGLE-3 draft head (Table~\ref{tab:speed-64}: $k=4$ without it).}
\label{tab:speed-async}
\centering
\input{tables/speed_async}
\end{minipage}

\end{table*}

\makeatletter
\begin{table*}[p]
\vspace{-6pt}%
\centering
\caption{\textbf{Real-kernel rollout engines at three batch sizes} (one run per cell). Left: our
training batch (64 sequences per step), steps 3--200, with the EAGLE-3 draft head (Table~\ref{tab:speed-64}: without
it). Right: 8-step runs without it at 1{,}024 and (bottom block) 1{,}536 sequences per step, where the BF16 cache
fills, steps 3--8. Budget: the quality verdict at 64 sequences per step. Tokens/h: generated tokens per hour. Peaks
are lower bounds. $^\dagger$Replica cache capacities differ by one 16-token block.}
\label{tab:speed-samplers}
\centering
\input{tables/speed_samplers}
\vspace{-3pt}%
\caption{\textbf{The training batch with and without the draft head} (64 sequences per step, one
run each). Updates per hour over steps 3--30, relative to BF16 in the same half. Left: 30-step runs without the
EAGLE-3 draft head, otherwise unchanged, added after the other results (App.~\ref{app:prereg:post}). Right: the
original runs, with it. Generation: seconds per step of generation and teacher scoring; for the asynchronous run, the
trainer's wait, and its ratio is to synchronous BF16.}
\label{tab:speed-64}
\centering
{\scriptsize\setlength{\tabcolsep}{3pt}%
\begin{tabular}{@{}lrrrr@{\hspace{12pt}}rr@{}}
\toprule
 & \multicolumn{4}{c}{Without the head} & \multicolumn{2}{c}{With the head} \\
\cmidrule(lr){2-5}\cmidrule(l){6-7}
 & Updates & $\times$ & Step & Generation & Updates & $\times$ \\
Rollout engine & per hour & BF16 & (s) & (s) & per hour & BF16 \\
\midrule
Standard engine (BF16) & 27.3 & 1 & 132.0 & 100.1 & 34.1 & 1 \\
vLLM's FP8 cache, first layer's KV cache in BF16 & 27.3 & 1.002 & 131.8 & 98.6 & 33.7 & 0.989 \\
Model Optimizer FP8, first and last layers in BF16 & 34.5 & 1.266 & 104.3 & 71.6 & 36.8 & 1.079 \\
Asynchronous, weights up to 4 versions old (BF16) & 54.8 & 2.008 & 65.7 & 0.6 & 56.0 & 1.643 \\
\bottomrule
\end{tabular}}

\vspace{-3pt}%
\begin{minipage}[t]{\columnwidth}\vspace{0pt}\centering
\caption{\textbf{Time to the same held-out KL} (eight GPUs each; one run each). At each checkpoint:
hours since the first training step, and held-out KL to the teacher (nats/token); step 0 is the untrained student.
The synchronous and $k=8$ runs are the repeats of App.~\ref{app:prereg:onerun} (a checkpoint every 25 steps); with
$k=8$ and FP8, training reaches the synchronous run's final value after 3.69 hours, 1.62 times sooner. Accuracy at
step 200: 61.54\% (synchronous) and 60.46\% ($k=8$ and FP8).}
\label{tab:speed-ttq}
\centering
\input{tables/speed_ttq}
\end{minipage}\hfill
\begin{minipage}[t]{\columnwidth}\vspace{0pt}\centering\def\@captype{figure}%
\includegraphics[width=\columnwidth]{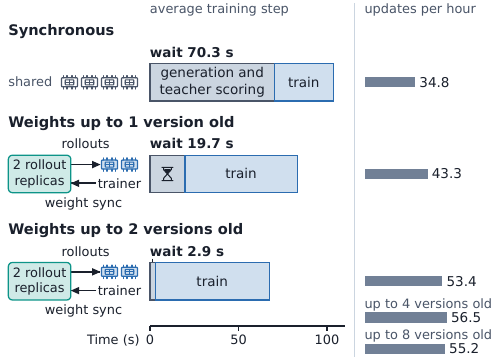}
\caption{\textbf{Where the trainer's step time goes} (eight GPUs, four for the teacher; one run per
configuration; with the draft head, Table~\ref{tab:speed-64} without it). One average step per row; trainer blocks to
scale; rollout side not timed. \textbf{Synchronous:} the trainer and four rollout replicas share four GPUs; the trainer
waits 70.3\,s of its 103.5\,s step. \textbf{Up to 1 or 2 versions old:} two GPUs train, two generate. \textbf{Right:}
updates per hour level off near 55.}
\label{fig:speed-steptime}\label{fig:diag-step-timeline}
\end{minipage}
\end{table*}
\makeatother

\makeatletter
\twocolumn[{\def\@captype{table}%
\centering
\caption{\textbf{Changing OPD's update} (one 200-step run per row; score centering, the advantage offset and
Eq.~\eqref{eq:app-drift}: App.~\ref{app:prereg:rules}). Displacement: the checkpoint minus the comparison run's, projected on
the noise direction $\hat n$; positive is toward the rollout engine. Held-out KL change: against each block's comparison run.}
\label{tab:mech-tests}
\centering
\input{tables/mech_tests}
\vspace{6pt}}]
\makeatother

%% file: tables/admission.tex
{\scriptsize\setlength{\tabcolsep}{3pt}%
\begin{tabular}{@{}p{1.95in}lrrrrrrll@{}}
\toprule
 & & \multicolumn{3}{c}{Mismatch (nats/token)} & \multicolumn{2}{c}{Held-out KL} & Accuracy & \multicolumn{2}{c}{Primary rule} \\
\cmidrule(lr){3-5}\cmidrule(lr){6-7}\cmidrule(l){9-10}
Rollout engine & Kernels & Estimate & Training & Ratio & Step 100 & Step 200 & (points) & Predicted & Observed \\
\midrule
\multicolumn{10}{@{}l}{\emph{7B student: NVIDIA Model Optimizer recipes (weights, activations and a calibrated FP8 attention cache)}} \\
FP8, every layer & real & 0.576 & 0.6465 & 1.12 & $+40.80\%$ & $+57.96\%$ & $-3.41$ & fail & fail \\
FP8, first and last layers \rev{in BF16} & real & 0.0054 & 0.0091 & 1.68$^\ast$ & $-1.60\%$ & $-0.45\%$ & $+0.88$ & pass & pass \\
NVFP4, every layer & emulated & 0.408 & 0.5702 & 1.40 & $+41.87\%$ & $+59.66\%$ & $-5.21^\dagger$ & fail & fail \\
NVFP4, first layer's KV cache in BF16 & emulated & 0.030 & 0.0467 & 1.56$^\ast$ & $-0.01\%$ & $-5.04\%$ & $+1.18$ & pass & pass \\
\midrule
\multicolumn{10}{@{}l}{\emph{7B student: vLLM's own FP8 attention cache at a fixed scale of 1 (BF16 weights and activations)}} \\
Every layer & real & 0.5205 & 0.6856 & 1.32 & $+39.55\%$ & $+53.92\%$ & $-3.63$ & fail$^a$ & fail \\
First layer's cache in BF16 & real & 0.0037 & 0.0058 & 1.58$^\ast$ & $-0.01\%$ & $-1.53\%$ & $+0.71$ & pass & pass \\
\midrule
\multicolumn{10}{@{}l}{\emph{7B student: asynchronous training at a staleness of 8 model versions, with the admitted FP8 recipe}} \\
FP8, first and last layers \rev{in BF16} & real & --$^b$ & 0.0127 & -- & $-1.79\%$ & $+0.45\%$ & -- & pass & pass \\
\midrule
\multicolumn{10}{@{}l}{\emph{Other pairs: Model Optimizer's FP8 recipe on every layer, and vLLM's own FP8 cache at scale 1 on every layer}} \\
1.5B student, Model Optimizer FP8 & real & 0.44$^c$ & 0.3913 & 0.89 & $+257.2\%$ & $+419.7\%$ & $-10.71^\dagger$ & fail & fail \\
1.5B student, vLLM FP8 cache & real & -- & 0.3645 & -- & $+268.3\%$ & $+478.5\%$ & $-10.87^\dagger$ & none & fail \\
1.5B student, vLLM FP8 cache, first layer's KV cache in BF16 & real & -- & 0.0030 & -- & $+1.0\%$ & $-4.3\%$ & $+0.53$ & pass$^e$ & pass \\
Llama 3B student, Model Optimizer FP8 & real & 0.0074$^c$ & 0.0098 & 1.32 & $+6.6\%$ & $+5.6\%$ & $-0.42$ & pass & pass \\
Llama 3B student, vLLM FP8 cache & real & -- & 0.0041 & -- & $-0.2\%$ & $+13.0\%$ & $-0.44$ & none & pass$^d$ \\
\bottomrule
\end{tabular}}

%% file: tables/speed_async.tex
{\scriptsize\setlength{\tabcolsep}{3pt}%
\begin{tabular}{@{}lrrrrrr@{}}
\toprule
 & Mismatch, & \multicolumn{2}{c}{Held-out KL (\%)} & Trainer & Updates & Ratio to \\
\cmidrule(lr){3-4}
Staleness $k$ & steps 1--100 & step 100 & step 200 & waits (s) & per hour & sync.\ run \\
\midrule
Synchronous & 0.00083 & (ref.) & (ref.) & 70.3 & 34.8 & 1 \\
1 & 0.00098 & $-1.93$ & $-2.72$ & 19.7 & 43.3 & 1.244 \\
2 & 0.00129 & $-1.35$ & $-0.82$ & 2.9 & 53.4 & 1.536 \\
4 & 0.00227 & $-1.21$ & $+1.41$ & 0.1 & 56.5 & 1.624 \\
8 & 0.00467 & $-0.95$ & $+0.29$ & 1.0 & 55.2 & 1.586 \\
\midrule
4, FP8 engine & 0.0104 & $-1.10$ & $-0.18$ & 0.1 & 56.6 & 1.629 \\
8, FP8 engine & 0.0127 & $-1.79$ & $+0.45$ & 0.3 & 53.8 & 1.547 \\
\bottomrule
\end{tabular}}

%% file: tables/speed_samplers.tex
{\scriptsize\setlength{\tabcolsep}{3pt}%
\begin{tabular}{@{}lrr@{\hspace{12pt}}rrrrrrr@{}}
\toprule
 & \multicolumn{2}{c}{64 sequences per step} & \multicolumn{7}{c}{8 steps at 1{,}024 sequences per step (256 per replica; bottom block 1{,}536, 384 per replica)} \\
\cmidrule(lr){2-3}\cmidrule(l){4-10}
 & Updates & $\times$ & Updates & $\times$ & Tokens/h & Capacity & Peak & Pre- & Mismatch \\
Rollout engine (real kernels) & per hour & standard & per hour & standard & $\times$ standard & (k tokens) & cache use & emptions & (steps 1--8) \\
\midrule
Standard engine (BF16) & 34.8 & 1 & 4.862 & 1 & 1 & 864 & 1.0000 & 261 & 0.0008 \\
\midrule
\multicolumn{10}{@{}l}{\emph{Inside the budget}} \\
vLLM's FP8 cache, first layer's KV cache in BF16 & 34.4 & 0.990 & 5.150 & 1.059 & 1.065 & 1{,}668 & 0.557 & 0 & 0.0044 \\
Model Optimizer FP8, first and last layers in BF16 & 37.6 & 1.082 & 5.428 & 1.116$^\dagger$ & 1.129 & 1{,}808 & 0.517 & 0 & 0.0065 \\
\midrule
\multicolumn{10}{@{}l}{\emph{Outside the budget}} \\
Model Optimizer FP8, every layer & 25.1 & 0.723 & 4.230 & 0.870 & 1.078 & 1{,}955 & 0.631 & 0 & 0.595 \\
vLLM's FP8 cache, every layer & 22.7 & 0.652 & \multicolumn{7}{c}{not run at this batch} \\
\midrule
\multicolumn{10}{@{}l}{\emph{8 steps at 1{,}536 sequences per step (384 per replica); the memory test of App.~\ref{app:speedbench}}} \\
Standard engine (BF16) & -- & -- & 3.234 & 1 & 1 & 863 & 1.0000 & 7{,}083 & 0.0008 \\
vLLM's FP8 cache, first layer's KV cache in BF16 & -- & -- & 3.594 & 1.111$^\dagger$ & 1.113 & 1{,}667 & 0.832 & 0 & 0.0043 \\
Model Optimizer FP8, first and last layers in BF16 & -- & -- & 3.755 & 1.161 & 1.171 & 1{,}808 & 0.760 & 0 & 0.0066 \\
\bottomrule
\end{tabular}}

%% file: tables/speed_ttq.tex
{\scriptsize\setlength{\tabcolsep}{3pt}%
\begin{tabular}{@{}rrrrrrr@{}}
\toprule
 & \multicolumn{2}{c}{Synchronous,} & \multicolumn{2}{c}{Asynchronous ($k=8$),} & \multicolumn{2}{c}{Asynchronous ($k=4$),} \\
 & \multicolumn{2}{c}{standard engine} & \multicolumn{2}{c}{FP8 engine} & \multicolumn{2}{c}{FP8 engine} \\
\cmidrule(lr){2-3}\cmidrule(lr){4-5}\cmidrule(l){6-7}
Step & Hours & Held-out KL & Hours & Held-out KL & Hours & Held-out KL \\
\midrule
0 & 0 & 0.096481 & 0 & 0.096481 & 0 & 0.096481 \\
25 & 0.767 & 0.081657 & 0.473 & 0.081714 & 0.449 & 0.080677 \\
50 & 1.555 & 0.066267 & 0.976 & 0.067087 & 0.943 & 0.067879 \\
75 & 2.282 & 0.061172 & 1.413 & 0.060787 & 1.361 & 0.061000 \\
100 & 3.050 & 0.056436 & 1.878 & 0.055715 & 1.810 & 0.056635 \\
125 & 3.783 & 0.054263 & 2.326 & 0.053763 & 2.234 & 0.053992 \\
150 & 4.541 & 0.051084 & 2.798 & 0.050712 & 2.688 & 0.051218 \\
175 & 5.251 & 0.049197 & 3.243 & 0.049157 & 3.110 & 0.048863 \\
200 & 5.980 & 0.047643 & 3.694 & 0.047602 & 3.537 & 0.048019 \\
\bottomrule
\end{tabular}}

%% file: tables/mech_tests.tex
{\footnotesize\setlength{\tabcolsep}{4pt}%
\begin{tabular}{@{}llrrrll@{}}
\toprule
 & & & \multicolumn{2}{c}{Displacement ($\times10^{-5}$)} & \multicolumn{2}{c}{Held-out KL change, \% [95\% CI]} \\
\cmidrule(lr){4-5}\cmidrule(l){6-7}
Update & Rollout engine & Mismatch & step 100 & step 200 & step 100 & step 200 \\
\midrule
\multicolumn{7}{@{}l}{\emph{Plain update, against the standard run}} \\
Plain & noise, $\sigma=0.0665$ & 0.0547 & \ensuremath{-43.45^{*}} & \ensuremath{-54.06^{*}} & \ensuremath{+5.27} [\ensuremath{+3.08}, \ensuremath{+7.50}] & \ensuremath{+6.35} [\ensuremath{+3.52}, \ensuremath{+9.15}] \\
Plain & noise, $\sigma=0.15$ & 0.0968 & \ensuremath{-70.02^{*}} & \ensuremath{-80.96^{*}} & \ensuremath{+0.48} [\ensuremath{-1.47}, \ensuremath{+2.52}] & \ensuremath{+0.26} [\ensuremath{-2.31}, \ensuremath{+2.85}] \\
Plain & noise, $\sigma=0.325$ & 0.3931 & \ensuremath{-33.21^{*}} & \ensuremath{-34.73^{*}} & \ensuremath{+19.54} [\ensuremath{+16.97}, \ensuremath{+22.11}] & \ensuremath{+23.28} [\ensuremath{+19.92}, \ensuremath{+26.60}] \\
\midrule
\multicolumn{7}{@{}l}{\emph{Score centering (drift removed), against the run with the same engine and the plain update}} \\
Centering & standard & 0.0008 & -- & -- & \ensuremath{-0.59} [\ensuremath{-2.99}, \ensuremath{+1.72}] & \ensuremath{-0.75} [\ensuremath{-3.34}, \ensuremath{+1.86}] \\
Centering & noise, $\sigma=0.0665$ & 0.0549 & \ensuremath{+6.46^{*\dagger}} & \ensuremath{+1.72^{\dagger}} & \ensuremath{+3.65} [\ensuremath{+1.20}, \ensuremath{+6.14}] & \ensuremath{+2.61} [\ensuremath{+0.03}, \ensuremath{+5.15}] \\
Centering & noise, $\sigma=0.15$ & 0.0970 & \ensuremath{+24.00^{*\dagger}} & \ensuremath{+24.70^{*\dagger}} & \ensuremath{+3.44} [\ensuremath{+1.05}, \ensuremath{+5.82}] & \ensuremath{+2.27} [\ensuremath{-0.35}, \ensuremath{+4.82}] \\
Control & standard & 0.0008 & -- & -- & \ensuremath{-0.25} [\ensuremath{-2.53}, \ensuremath{+2.01}] & \ensuremath{-0.98} [\ensuremath{-3.47}, \ensuremath{+1.54}] \\
\midrule
\multicolumn{7}{@{}l}{\emph{Advantage offset $b$ (drift coefficient shifted by $b$), against the standard run}} \\
$b=+0.15$ & noise, $\sigma=0.15$ & 0.0975 & \ensuremath{+2.38^{*}} & \ensuremath{+95.46^{*}} & \ensuremath{-2.88} [\ensuremath{-4.90}, \ensuremath{-0.86}] & \ensuremath{-6.49} [\ensuremath{-9.24}, \ensuremath{-3.78}] \\
$b=+0.5$ & noise, $\sigma=0.15$ & 0.0928 & \ensuremath{+132.1^{*}} & \ensuremath{+409.7^{*}} & \ensuremath{-4.38} [\ensuremath{-6.43}, \ensuremath{-2.32}] & \ensuremath{-6.86} [\ensuremath{-9.49}, \ensuremath{-4.29}] \\
$b=-0.15$ & noise, $\sigma=0.15$ & 0.0950 & \ensuremath{-75.02^{*\ddagger}} & \ensuremath{-179.6^{*\ddagger}} & \ensuremath{+4.30} [\ensuremath{+2.21}, \ensuremath{+6.45}] & \ensuremath{+5.66} [\ensuremath{+3.04}, \ensuremath{+8.30}] \\
$b=+0.15$ & standard & 0.0008 & \ensuremath{-0.69} & \ensuremath{-1.78} & \ensuremath{-1.63} [\ensuremath{-3.53}, \ensuremath{+0.30}] & \ensuremath{+0.32} [\ensuremath{-2.31}, \ensuremath{+2.96}] \\
$b=+0.5$ & standard & 0.0008 & \ensuremath{+0.34} & \ensuremath{-0.05} & \ensuremath{-1.37} [\ensuremath{-3.18}, \ensuremath{+0.43}] & \ensuremath{+0.61} [\ensuremath{-2.16}, \ensuremath{+3.26}] \\
\midrule
\multicolumn{7}{@{}l}{\emph{Truncated importance sampling (token weights capped at 2), against the run on the same engine without it}} \\
Truncated & noise, $\sigma=0.15$ & 0.0960 & -- & -- & \ensuremath{+2.90} [\ensuremath{+0.45}, \ensuremath{+5.30}] & \ensuremath{+6.97} [\ensuremath{+4.54}, \ensuremath{+9.41}] \\
Truncated & FP8, every layer & 0.6284 & -- & -- & \ensuremath{-1.86} [\ensuremath{-3.97}, \ensuremath{+0.21}] & \ensuremath{-1.23} [\ensuremath{-3.27}, \ensuremath{+0.78}] \\
\bottomrule
\multicolumn{7}{@{}p{0.86\textwidth}@{}}{\scriptsize\rule{0pt}{2.2ex}%
$^{*}$~Beyond the 97.5th percentile of the same projection on 200 random directions.
$^{\dagger}$~Score-centered run minus the plain run with the same engine.
$^{\ddagger}$~Against the run at $\sigma=0.15$ without offset.
Intervals: 95\%, problem-paired bootstrap. The standard run's reference at step 100 is the mean of three
earlier 100-step standard runs, and at step 200 (and for displacements) the 200-step run. Mismatch: mean over steps
1--200, nats/token. Control: the score-centering loss without the centering term. FP8, every layer: Model
Optimizer's FP8 recipe on every layer.} \\
\end{tabular}}

%% file: sections/a-limitations.tex
\section{Limitations}
\label{app:limitations}\label{sec:limitations}
We study reverse-KL OPD with the single-sample estimator of Eq.~\eqref{eq:bg-adv}, on math.
\begin{itemize}
\setlength{\itemsep}{0pt}\setlength{\parskip}{0pt}
\item \emph{One run per configuration.} Intervals resample problems, not runs, so verdicts are provisional;
  repeated configurations differ by 1.0\% and 1.5\% in held-out KL (App.~\ref{app:prereg:onerun}). The second runs at
  0.097 and 0.39, and most likely the FP8 run with weights up to 4 versions old, used a newer GPU driver without a
  control run, so that run's speed ratios include any driver difference (App.~\ref{app:repro}).
\item \emph{One main pair.} The budget, 0.184 to 0.39~nats/token, is our 7B student's, and no run lies
  inside it. Our teacher is an RL fine-tune of the student; a pair farther apart may have a smaller budget. The 1.5B
  student's budget is not located, one Llama-3.2-3B run detects only held-out KL rises of 14 to 20\% or more, and no
  training result covers Qwen3, whose 4B student failed our check that plain OPD trains on it.
\item \emph{Length, hardware and speed.} Runs last at most 200 steps on GH200 GPUs; the noise level whose
  verdict changed between steps 100 and 200 moved toward failure in both runs. NVFP4 is emulated (quality only). Each
  FP8 speed gain comes from one short run, and the memory saving's credit, 5.7\%, just passes its 5\% bar.
\item \emph{Time to quality and admission.} With FP8, weights up to 8 versions old reach the synchronous
  run's final held-out KL only at the last checkpoint; up to 4 versions old, they end 0.8\% above it, within the spread
  of repeated runs (Table~\ref{tab:speed-ttq}). No admission prediction was tested near the budget's edge
  (App.~\ref{app:prereg:admit}).
\item \emph{Staleness and RL.} Stale-weight runs stay below 0.005~nats/token, so OPD's staleness limit is
  not located; GRPO and REINFORCE did not pass our learning test in 200 steps.
\item \emph{Mechanism.} The fixed prefixes may overstate how much the error shrinks in training: there the
  KL to the teacher falls 28-fold, against 2-fold for held-out KL. Still open: why noise at 0.39 fails only by step 200,
  why per-token corrections did not help, and why positive offsets lower held-out KL; the larger offset costs 2.9
  points of accuracy [$-4.95$, $-0.91$].
\end{itemize}